%% file: main.tex
\documentclass[10pt,twocolumn,letterpaper]{article}

\usepackage[pagenumbers]{iccv} 

\input{preamble}

\definecolor{iccvblue}{rgb}{0.21,0.49,0.74}
\usepackage[pagebackref,breaklinks,colorlinks,allcolors=iccvblue]{hyperref}
\usepackage{xspace}
\usepackage{bbding}
\usepackage{multirow}
\usepackage{dsfont}
\usepackage[hypcap=false]{caption}
\newcommand{\mname}{\textcolor{black}{HPSv3}\xspace}
\newcommand{\dname}{\textcolor{black}{HPDv3}\xspace}

\def\paperID{2872} 
\def\confName{ICCV}
\def\confYear{2025}

\title{HPSv3: Towards Wide-Spectrum Human Preference Score}

\author{
Yuhang Ma$^{1,3*}$ \qquad
Yunhao Shui$^{1,4*}$ \qquad
Xiaoshi Wu$^{2}$ \qquad
Keqiang Sun$^{1,2\dagger}$ \qquad
Hongsheng Li$^{2,5,6\dagger}$ \\
\centerline{$^{1}$~Mizzen AI \qquad $^{2}$~CUHK MMLab \qquad $^{3}$~King's College London}\\
\centerline{$^{4}$~Shanghai Jiaotong University \qquad $^{5}$~Shanghai AI Laboratory \qquad $^{6}$~CPII, InnoHK} \\
\centerline{
\texttt{\small \{yhma,kqsun,yhshui\}@mizzen.ai}, 
\texttt{\small \{wuxiaoshi@link, hsli@ee\}.cuhk.edu.hk}}
}

\newcommand\blfootnote[1]{%
  \begingroup
  \renewcommand\thefootnote{}\footnote{#1}%
  \addtocounter{footnote}{-1}%
  \endgroup
}
\begin{document}

\twocolumn[
\maketitle
\input{figure/teaser_main}
\bigbreak
]

\blfootnote{$*$Equal contribution, $\dagger$Equal advising}

\input{sec/abstract}
\input{sec/introduction}

\input{sec/relatedwork}

\input{sec/method}

\input{sec/experiment}

\input{sec/conclusion}
\newpage
\input{sec/acknowledgement}

{
    \small
    \bibliographystyle{ieeenat_fullname}
    \bibliography{main}
}

\clearpage
\maketitlesupplementary
\setcounter{page}{1}
\setcounter{figure}{0}
\setcounter{section}{0}
\renewcommand{\thefigure}{S\arabic{figure}}
\setcounter{table}{0}
\renewcommand{\thetable}{S\arabic{table}}
\input{sec/X_suppl}

\end{document}

%% file: preamble.tex
\usepackage{graphicx}

%% file: figure/teaser_main.tex
\begin{center}
\vspace{-15pt}
    \includegraphics[width=0.85\linewidth]{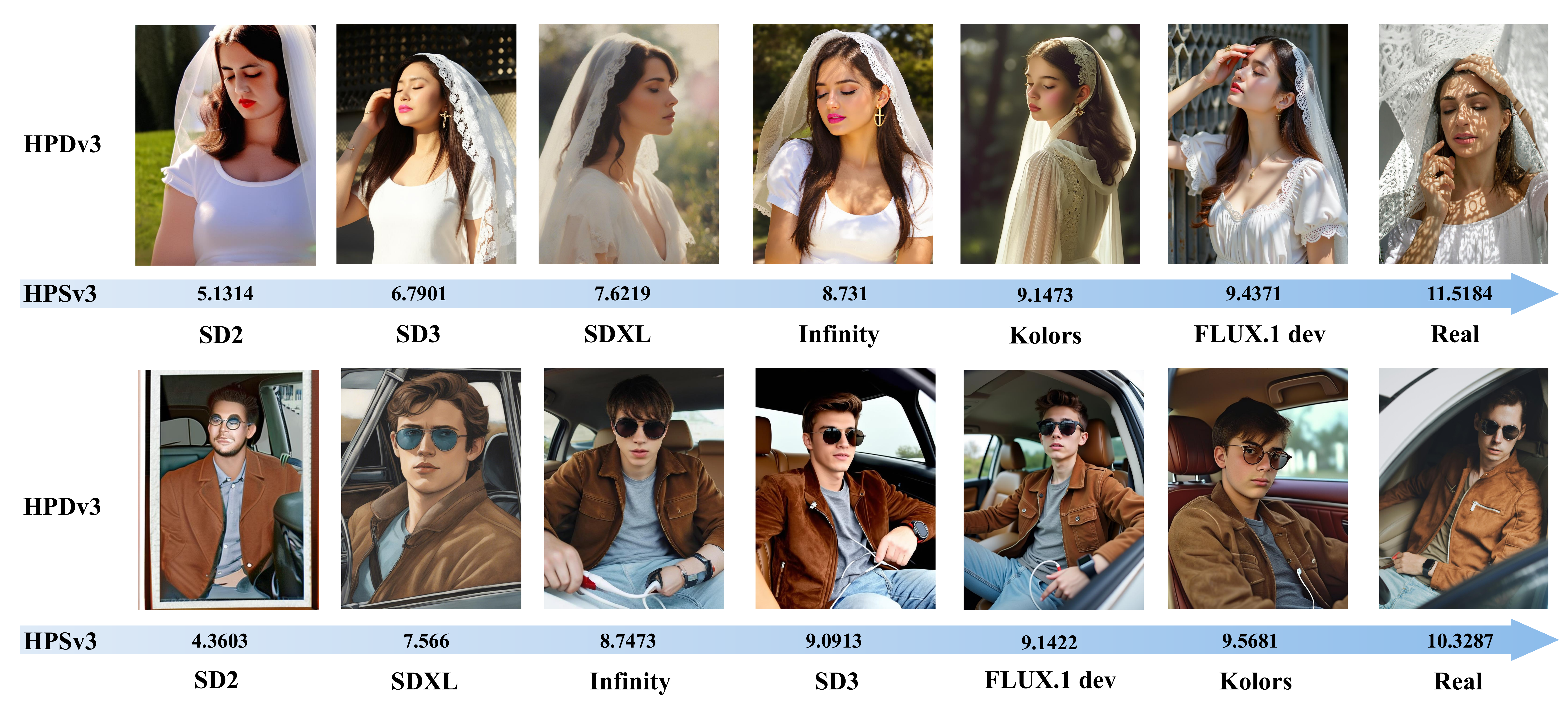}
    
\end{center}
\vspace{-10pt}
\captionof{figure}{\textbf{Wide-Spectrum Human Preference Score (HPSv3)}. Examples from the HPDv3 dataset with their corresponding HPSv3 scores. HPDv3 represents the first comprehensive wide-spectrum human preference dataset, designed to evaluate generative models across a wide range of prompts and scenarios.} 
\label{fig:teaser}
\vspace{-0.5em}

%% file: sec/abstract.tex
\begin{abstract}

Evaluating text-to-image generation models requires alignment with human perception, yet existing human-centric metrics are constrained by limited data coverage, suboptimal feature extraction, and inefficient loss functions. To address these challenges, we introduce Human Preference Score v3 (\mname). (1) We release \dname, the first wide-spectrum human preference dataset integrating 1.08M text-image pairs and 1.17M annotated pairwise comparisons from state-of-the-art generative models and low to high-quality real-world images. (2) We introduce a VLM-based preference model trained using an uncertainty-aware ranking loss for fine-grained ranking.
Besides, we propose Chain-of-Human-Preference (CoHP), an iterative image refinement method that enhances quality without extra data, using \mname to select the best image at each step.
Extensive experiments demonstrate that \mname serves as a robust metric for wide-spectrum image evaluation, and CoHP offers an efficient and human-aligned approach to improve image generation quality. The code and dataset are available at the \href{https://mizzenai.github.io/HPSv3.project/}{HPSv3 Homepage}.
\end{abstract}

%% file: sec/introduction.tex
\section{Introduction}

With the rapid advancement of text-to-image generation models, human-centric evaluation methods, such as HPS~\cite{wu2023hps,hpsv2}, ImageReward \cite{xu2023imagereward}, PickScore \cite{kirstain2023pick}, and MPS~\cite{mps}, have emerged to incorporate subjective human feedback. These methods construct human-annotated datasets and train models to capture the distribution of human preferences, enabling more perceptually aligned evaluations.

However, these human-centric evaluation models remain limited by constrained data distributions, suboptimal model designs, and simplified training schemes.

On the one hand, prior human-centric evaluation metrics~\cite{wu2023hps, kirstain2023pick, xu2023imagereward} are limited by the narrow quality spectrum of their training data, which primarily consists of images generated by diffusion-based models~\cite{stable_diffusion}. This makes them insufficient for evaluating more advanced approaches, such as diffusion transformers~\cite{chen2023pixart, flux2024} and autoregressive models~\cite{tian2024var, han2024infinity}. In addition, the lack of standardized and reliable annotation practices introduces potential biases in the data labeling process.

On the other hand, from a model architecture perspective, commonly used feature extractors such as CLIP~\cite{radford2021clip} and BLIP~\cite{li2022blip}, employed in works like~\cite{wu2023hps, hpsv2, kirstain2023pick, xu2024imagereward}, have been shown to be less effective at multimodal feature extraction~\cite{blip2,qwen2vl}, resulting in less comprehensive representations for final ranking. Moreover, directly optimizing models with KL divergence~\cite{hpsv2, mps} overlooks potential inconsistencies in annotation ground truth, a common issue in human preference annotation. These challenges highlight the need for a more robust model architecture and training paradigm that better captures human preferences.

This paper introduces two key advancements to solve these challenges: (a) a meticulously annotated dataset spanning the wide quality spectrum, enriched with diverse text-image pairs and curated through a carefully designed pipeline to align with human preferences, and (b) a preference model with a robust backbone for comprehensive feature encoding, paired with an uncertainty-aware ranking loss to alleviate the uncertainty or errors in the training samples caused by annotators.

As illustrated in \Cref{fig:teaser}, inspired by the concept of the spectrum~\cite{Newton1672Spectrum}, we introduce Human Preference Dataset v3 (\dname) — a more comprehensive and systematic dataset comprising \textbf{1.08 million} text-image pairs and \textbf{1.17 million} annotated pairwise comparisons. \dname\ expands upon previous efforts by integrating outputs from state-of-the-art autoregressive and diffusion models alongside high-quality real images sourced from the Internet, representing the upper bound of human-collected image quality.

Specifically, \dname\ is built from three key sources: (1) expanding HPDv2 by generating images with $10$ additional state-of-the-art models, (2) curating high-quality Internet photographs, captioned by Visual Language Models (VLMs) to generate additional images, and (3) incorporating Midjourney outputs, including four generated images per prompt and a user-selected preference.
To enhance reliability, we conduct pairwise comparisons with $9\sim19$ human annotators per sample, ensuring diverse subjective evaluations. \dname\ mitigates biases in prior datasets, serving as both a foundation for training preference-aligned models and a benchmark for evaluating generative models.

Based on \dname, we introduce Human Preference Score v3 (\mname), a comprehensive model trained on the \dname dataset, to better align image generation models with human expectations. \mname\ features two key advancements: (1) it leverages powerful visual encoders from Vision-Language Models (VLMs) to capture rich, high-dimensional features, and (2) it employs an uncertainty-aware ranking loss to more accurately distinguish subtle differences in image quality. Trained on a diverse dataset of both generated and high-quality real images, \mname\ offers a robust and reliable evaluation framework that closely mirrors human preferences.

Finally, we introduce Chain-of-Human-Preference (CoHP), a novel, reasoning-driven process for image generation. CoHP's first innovation is the integration of Chain-of-Thought (CoT) inspired reasoning, which systematically improves image quality without the need for a training dataset. Second, it utilizes \mname\ as a reward model to evaluate images based on semantic alignment, realism, and aesthetic appeal. By guiding the reasoning process, \mname\ ensures the selection of the highest-quality image at each step, leading to a progressively refined and enhanced final output.

The contribution of \mname are three folded:
\begin{itemize}
    \item We propose the wide-spectrum human preference dataset \dname by integrating high-quality real-world images and state-of-the-art generative model outputs, including 1.08M text-image pairs and 1.17M annotated pairwise comparisons. This serves as a nuanced benchmark for evaluating generative models.
    \item We introduce \mname, a human preference model trained with \dname, which levearages the feature of VLMs and is trained using an uncertainty-aware ranking loss to discern the uncertainty or errors in training samples caused by annotators.
    \item We introduce CoHP, a novel reasoning approach to enhance image generation quality by iteratively refining outputs using \mname. 
\end{itemize}

%% file: sec/relatedwork.tex
\section{Related Work}
\subsection{Text-to-image Generation}
Generative models have significantly advanced from the rise of GANs~\cite{sun2022cgof, sauer2023stylegan, sun2023cgof++, piao2021inverting} to diffusion models ~\cite{stable_diffusion, dalle2, imagen, li2024ecnet, liu2024llm4genleveragingsemanticrepresentation, sun2024genca, laiinstantportrait, wu2024deepreward}. Recently, text-to-image models like FLUX~\cite{flux2024,hunyuandit,sd3} have further contributed to the field by integrating large-scale transformer architectures into diffusion models, enhancing the quality and coherence of generated images. Despite diffusion-based models, visual auto-regressive models~\cite{han2024infinity, fan2024fluid, wu2025dcar} also show strong capable of generating high-resolution, photorealistic images following language instruction. 

\input{tables/datasetcomparision}

\subsection{Human Preference Models}

 Recently, several studies such as HPS~\cite{wu2023hps}, HPSv2~\cite{hpsv2}, MPS~\cite{mps} and PickScore~\cite{kirstain2023pick} have suggested fine-tuning CLIP based on human preferences for images generated from the same textual prompt. ImageReward~\cite{xu2023imagereward} utilizes BLIP encoder as the feature extractor to train a reward model to evaluate and improve the generative models. However, these models have primarily been adjusted using images from Stable Diffusion and its variants, indicating that their ability to generalize across different datasets has yet to be validated. Also, utilizing CLIP/BLIP encoders as feature extractors is constrained due to either broad semantic information or limited feature dimensions, which may not effectively encapsulate user behaviors and expectations.

\subsection{Image quality dataset}

Text-to-image generative models often rely on high-quality real images sourced from the Internet to ensure the fidelity of their generated outputs. HPDv1~\cite{wu2023better}, ImageReward~\cite{xu2023imagereward} and Pick-a-Pic~\cite{kirstain2023pick} are human-annotated datasets of comparison between paired images, focusing on evaluating generated images and improving the image quality by aligning text-to-image models with human preference. Similar to these dataset, HPDv2~\cite{hpsv2} includes more images from a wider range of generative models, enabling a more comprehensive evaluation of these preference prediction models.
However, these datasets mainly compare images generated from existing models, neglecting comparisons with high-quality real images.

%% file: tables/datasetcomparision.tex
\begin{table*}[!t]
\centering
\small

\scalebox{0.72}{
\setlength{\tabcolsep}{0.7mm}{
\begin{tabular}{l|cc|ccccccc|cccc|cc|ccc|cccc|ccc}
\toprule
\multirow{2}{*}{Benchmark} && \multirow{2}{*}{Venue} && \multicolumn{5}{c}{\textbf{Image Source}} &&  \multicolumn{3}{c}{\textbf{Image Nums}} && \multicolumn{1}{c}{\textbf{Pair Nums}} && \multicolumn{2}{c}{\textbf{Prompt Source}} && \multicolumn{3}{c}{\textbf{Annotator}} && \multicolumn{2}{c}{\textbf{Choices}}    \\
& & & & HQI & LQI& GAN & Diffusion & AR & & Real & Generated & Total& & Total & & \centering User & Generated && \centering Source & Num/image & Convergency && 1 of 4 & 1 of 2 \\
\midrule
     HPDv1~\cite{wu2023better} && ICCV'23 && \XSolidBrush & \XSolidBrush & \XSolidBrush& \Checkmark& \XSolidBrush&& - & 99k & 99k && 25k && \Checkmark &\XSolidBrush && Users &  -  & - && \Checkmark &\XSolidBrush \\
     ImageRewardDB~\cite{xu2023imagereward} && NeurIPS'23 && \XSolidBrush &\XSolidBrush & \XSolidBrush& \Checkmark& \XSolidBrush&& - & 50k & 50k && 137k && \Checkmark &\XSolidBrush && Annotators & -  & - && \XSolidBrush &\Checkmark \\
     Pick-a-Pic~\cite{kirstain2023pick} && NeurIPS'23 && \XSolidBrush &\XSolidBrush & \XSolidBrush& \Checkmark& \XSolidBrush&& - & 638k & 638k && 584k && \Checkmark &\XSolidBrush && Users & -  & - && \XSolidBrush &\Checkmark \\
     
     HPDv2~\cite{hpsv2} && arXiv'23 && \XSolidBrush & \Checkmark &\Checkmark& \Checkmark& \Checkmark && 28k & 430k & 458k && 798k && \Checkmark &\XSolidBrush && Annotators & 10 persons & 59.9\% && \XSolidBrush &\Checkmark \\
     MHP~\cite{mps} && CVPR'24&& \XSolidBrush & \XSolidBrush &\Checkmark& \Checkmark& \Checkmark && - & 608k & 608k && 918k && \Checkmark &\Checkmark && Annotators & - & - && \XSolidBrush &\Checkmark \\
     HPDv3(ours) && ICCV'25 && \Checkmark & \Checkmark & \Checkmark & \Checkmark& \Checkmark&& \textbf{58k} & \textbf{1.03M} & \textbf{1.08M} && \textbf{1.17M} && \Checkmark &\Checkmark && Annotators & \textbf{9-19} persons & \textbf{76.5\%} && \Checkmark &\Checkmark\\
     
\bottomrule
\end{tabular}
}}
\caption{\label{tab:dataset_comparison}\textbf{Comparison with related datasets.} HPDv3 is the only dataset that comprehensively covers all types of image generation models and includes both high-quality (HQI) and low-quality (LQI) images across diverse prompts, featuring the largest image collection to date. Its annotations exhibit strong inter-rater agreement, ensuring exceptional reliability.}
\vspace{-5mm}
\end{table*}

%% file: sec/method.tex
\section{Human Preference Dataset v3}

Existing human preference benchmarks fail to provide a robust evaluation framework for generative models due to two key limitations: the absence of high-quality real photographs to define an upper quality bound, and the use of outdated generative models.

We present Human Preference Dataset v3 (\dname) to overcome these issues. This comprehensive dataset contains $1.17$ million binary preference choices across $1.08$ million images, grouped in pairs by prompt. The annotations are highly reliable, with each decision validated by $9$ to $19$ specialist annotators to meet a $90\%$ agreement threshold. Source model details are provided in the supplementary materials.

\subsection{Text-image Pairs Collection}
HPDv2 represents the most extensive annotated human preference dataset to date, comprising $798,000$ binary preference choices over $434,000$ images. However, its utility is constrained by two primary deficiencies. First, its image corpus consists of outputs from generative models and low-quality real images, thereby excluding the high-quality, real-world photographs, which is essential for establishing a robust upper quality bound. Second, the dataset's model coverage is outdated, extending only to Stable Diffusion 2.0 and consequently lacking comparisons with more recent state-of-the-art models.

\dname addresses these limitations by augmenting the HPDv2 dataset with these missing components: high-fidelity real photographs and images from current leading generative models. Specifically, the collection of \dname is comprised of three main components:

\noindent\textbf{Extending HPDv2.} We retain prompts from HPDv2, which include $103,700$ text entries refined by ChatGPT from sources like COCO Captions and DiffusionDB. We incorporate outputs from recently released state-of-the-art image generation models, such as FLUX.1-dev~\cite{flux2024}, Infinity~\cite{han2024infinity}, Hunyuan~\cite{hunyuandit}, Kolors~\cite{kolors}, and SD3~\cite{sd3} using these text entries. 

\noindent\textbf{Generation based on real photographs' captions.}
We incorporate high-quality real-world photographic images alongside AI-generated content, to build a wide-spectrum human preference dataset.
\begin{itemize}
    \item Prompt Categorization. To authentically align with the distribution of user prompts, we categorize the prompts into $12$ distinct categories.
    \item Distribution Alignment. We source high-quality photographic images from the Internet according to these categories, and ensure the portion of each category following the prompt distribution of JourneyDB~\cite{journeydb}.
    \item Aesthetic Filtering. To ensure quality, the Aesthetic Predictor is applied to score the collected data. As some categories like products inherently have lower aesthetic scores, the images with the top $10\%$ aesthetic score are selected for each category, resulting in a collection of $57,759$ high-quality images.
    \item Captioning and Generation. Visual language models are used to generate descriptions for these images, creating a prompt list, with which various generative models produce corresponding images.
\end{itemize}

\noindent\textbf{Collection from Midjourney.} In addition, Midjourney~\footnote{https://www.midjourney.com/}, a text-to-image generation model favored by artists, provides a wealth of user-written prompts, generated images, and user preference outcomes on the Discord platform. We collect \textbf{331,955} user-generated images and make them into pairwise data. These selection results can be directly used as authentic user labels.

\subsection{Annotation Pipeline}

We conduct the annotation by inviting users to select a preferred image from an image pair with the identical text prompt. Our annotation emphasis the annotator expertise and annotation quality. 

To ensure annotator proficiency, we created a validation set comprising $600$ image pairs annotated by $20$ professional artists, achieving an $80\%$ convergence rate. This set assesses annotator capability. Before participating, annotators must correctly evaluate at least $16$ out of $20$ randomly selected pairs from this validation set. For annotation quality control, each image pair in the main dataset is evaluated by $9$ to $19$ annotators, ensuring reliability and consistency. Annotators select images based on aesthetics, semantic similarity to the prompt, and overall coherence. Recognizing the inherent subjectivity of preference, no fixed acceptance standard is imposed. This annotation methodology was applied to the first two components of \dname. Image pairs with an inter-annotator confidence level exceeding $95\%$ were used for \mname training. For Midjourney data, we utilized preference labels collected directly from Discord. In total, we gathered $1.17$ million high-confidence pairwise comparisons for training.

\subsection{Comparison with Other Datasets}
The comparison of \dname with other existing datasets, as shown in \Cref{tab:dataset_comparison}, highlights several key advantages of the HPDv3 dataset.

Firstly, HPDv3 features a comprehensive range of model-generated results, utilizing $16$ different models including GAN-based, difffusion-based and autoregressive models. This surpasses the model diversity found in other datasets and ensures a wider variety of image generation styles and capabilities, providing the largest dataset of both text-image pairs and pairwise samples. 

Secondly, HPDv3 offers diverse prompt sources and image categories. Unlike other datasets that rely on limited sources like COCO~\cite{cococaptions} or DiffusionDB prompts, HPDv3 incorporates user-written prompts from JourneyDB, Midjourney and image captions from high-quality photography. This diversity enhances the dataset's applicability across various scenarios and supports more robust research.

Lastly, HPDv3 uniquely includes comparisons between real and generated images by incorporating both real high-quality images and low-quality images. This allows for a more nuanced comparison, helping researchers better understand the strengths and limitations of image-generation models relative to actual photographs. 

We also conduct an analysis of the annotation convergence between HPDv2 and \dname. The average convergence for HPDv2 is $59.9\%$. In contrast, \dname exhibits a higher concentration of convergence, resulting in the average convergence of $76.5\%$. This clearly indicates that \dname outperforms HPDv2 in dataset construction and is more effective in reaching high agreement in annotations.

In summary, \dname represents the largest and most comprehensive image generation preference dataset, uniquely covering all major model types, diverse prompt sources, and extensive choice typologies. The HPDv3 annotations exhibit high confidence and reliability, validated by strong annotator convergence.

\section{Human Preference Score v3}

While human evaluation remains the gold standard for assessing preferences toward synthetic images, it suffers from high costs and limited scalability. To address this, we introduce Human Preference Score v3 (\mname), a model that learns human preferences from annotations to serve as a scalable and automated evaluator.

\subsection{Model Design}

\paragraph{Architecture.} 
Vision-language models (VLMs) are widely used in downstream tasks such as image classification and tagging, owing to their powerful representational capabilities. Inspired by this, we use QWen2-VL~\cite{qwen2vl} as the backbone to extract features from both images and text. These features are then processed through a multilayer perceptron (MLP) to map and produce the final output.
In our approach, for a given pair of training images $(x_1, x_2)$, along with their corresponding text prompt $c$ and human preference annotation $(y_1, y_2)$, we derive the final outputs of the pair data using the following equations:
\begin{equation}
    r_1 = f_\phi(\mathcal{E}_{\theta}(x_1,c)),
    r_2 = f_\phi(\mathcal{E}_{\theta}(x_2,c)).
\end{equation}
Here, $\mathcal{E}_{\theta}$ denotes the vision-language model, $f_\phi$ denotes the multilayer perceptron (MLP).
\begin{figure}[t]
  \centering
  \includegraphics[width=\linewidth]{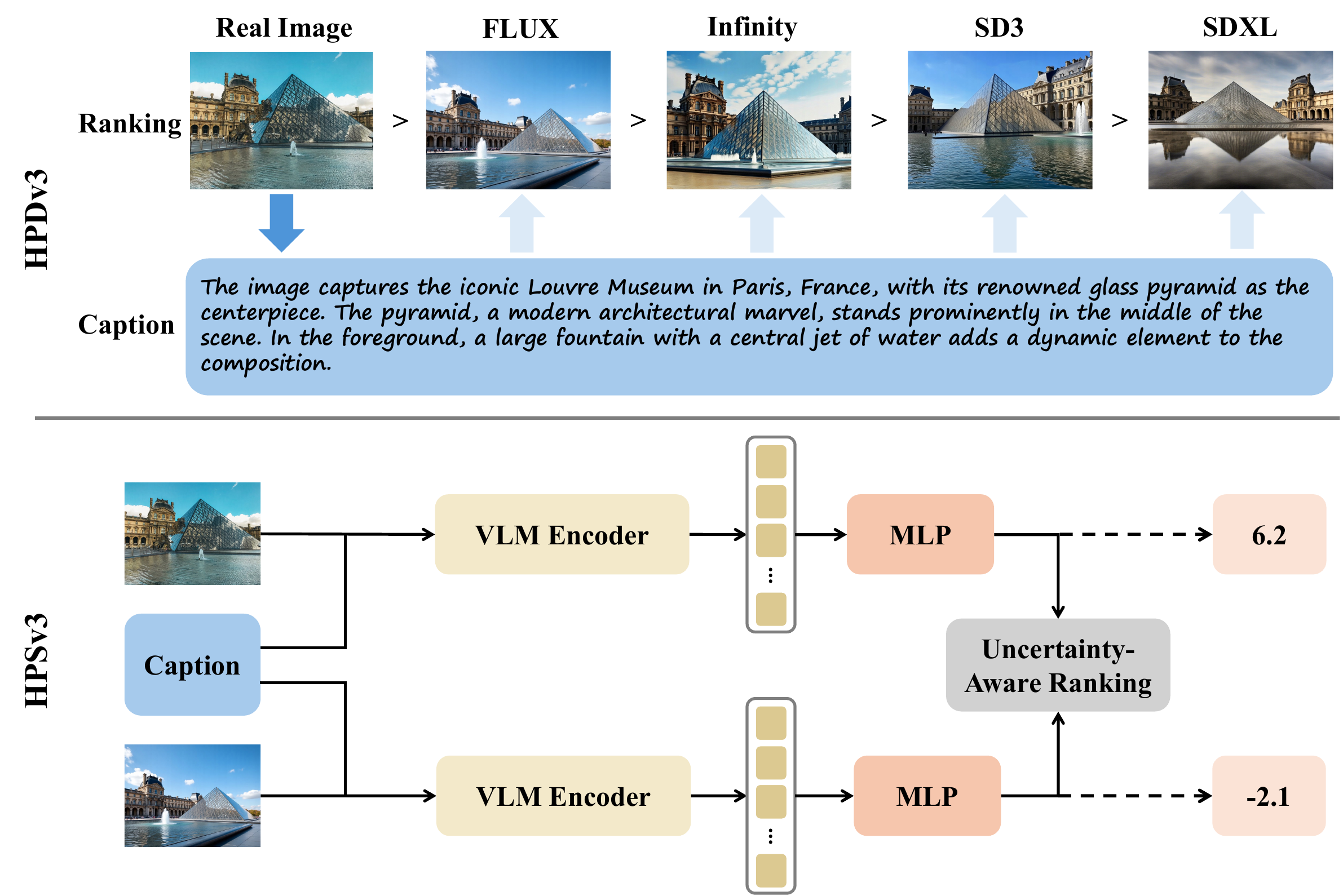}
  \caption{\textbf{An overview of HPDv3 and HPSv3.} HPDv3 integrates both real-world collected and generated images. HPSv3 employs a VLM backbone to extract rich semantic representations from images and captions, then utilizes uncertainty-aware ranking to effectively learn human preferences from paired comparison data.}
  \label{fig:framework}
\end{figure}
\paragraph{Uncertainty-Aware Ranking.} 
Prior models do not account for potential inconsistencies in annotation results. Directly selecting a certain preference score may introduce bias in the model's judgment of hard cases. 

Given the input $\mathcal{E}_{\theta}(x,c)$ and the label $y$, prior models compute a score $r$, and the preference probability is defined as:

\begin{equation}
    P(x_1 \succ x_2|c)=\text{sigmoid}(r_1-r_2),
\end{equation}

where the sigmoid function can be expressed as:
$
    \text{sigmoid}(x) = {1}/(1 + e^{-x}),
    \label{eq:sigmoid}
$
to constrain the predicted output to a probability value within the [0,1] interval.

Howerver, This deterministic modeling approach assigns equal confidence to all output predictions. During the training process, the model blindly assigns scores to samples without considering the uncertainty of predictions. 

Inspiring by \cite{uncertainty}, we introduce an uncertainty-aware ranking model. Unlike the traditional RankNet model, which utilizes the final linear layer of $f_\phi$ to output the score $r$, our approach uses the last two linear layers to predict $\mu$ and $\sigma$, modeling the output score $r$ as a one-dimension Gaussian distribution $r \sim \mathcal{N}(\mu, \sigma)$. This introduces an uncertainty aspect to the output score, alleviating the uncertainty or errors in pairwise data caused by the uncertainty of annotator labeling. The final probability is defined as:

\vspace{-0.5cm}
\begin{equation}
\begin{split}
P(x_1 \succ x_2|c) &= \iint \text{sigmoid}(r_1-r_2) \mathcal{N}(r_1 \mid \mu_1, \sigma_1) \\
&\quad \times \mathcal{N}(r_2 \mid \mu_2, \sigma_2) \, \mathrm{d}r_1\mathrm{d}r_2, \\
\end{split}
\end{equation}
\vspace{-0.5cm}

\begin{figure}[!t]
    \centering
    \includegraphics[width=1.02\linewidth]{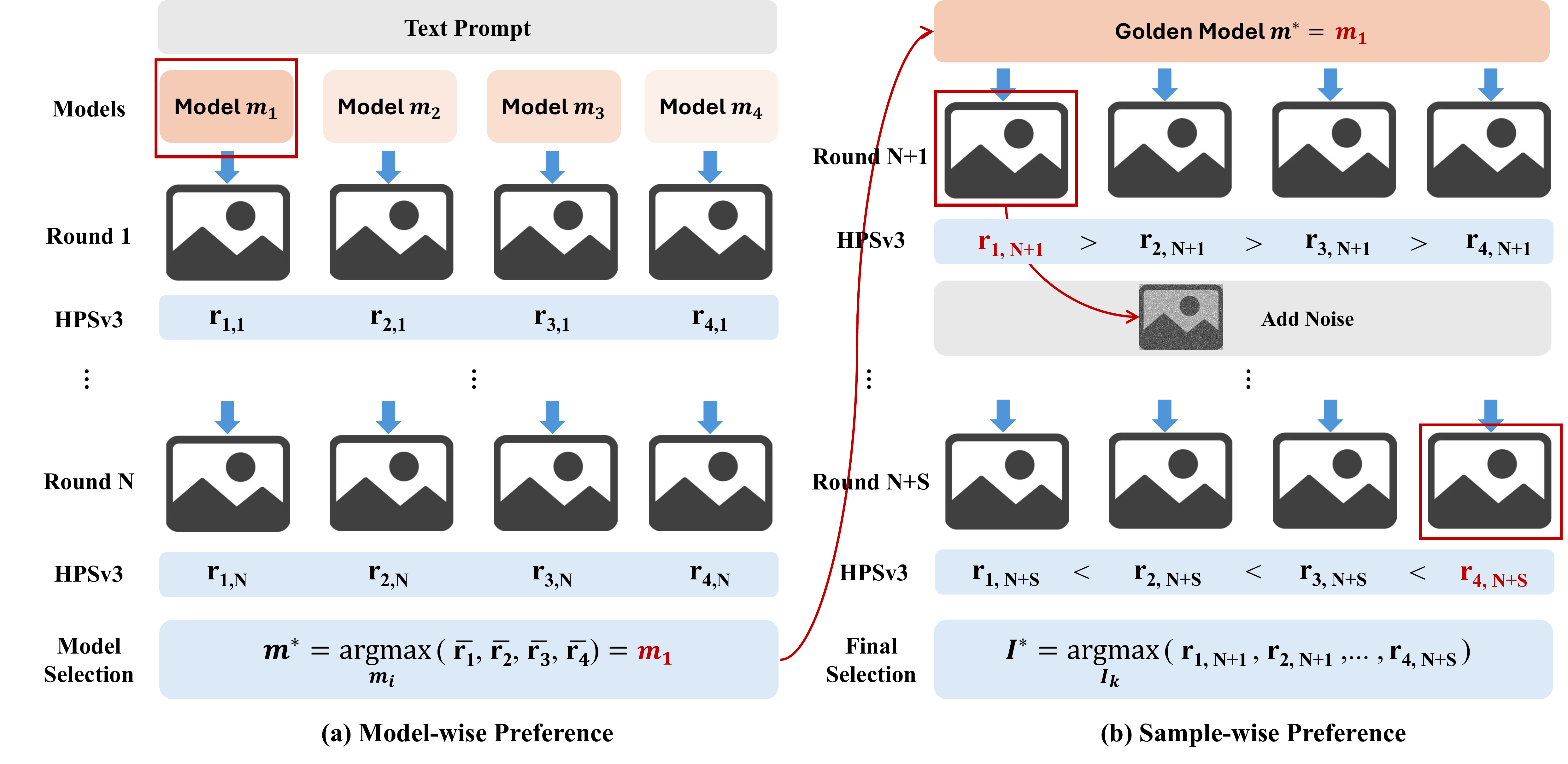}
    \caption{\textbf{The overview of Chain-of-Thought rounded image generation with \mname (CoHP).} CoHP incorporates both model-wise and sample-wise preferences, selected by \mname, to build a thinking-and-choosing image generation process.}
    \vspace{-4mm}
    \label{fig:cohp}
\end{figure}
\noindent We denote the higher-preferred rank sample as \( x_h \) with its corresponding reward \( r_h \), and the lower-preferred rank sample as \( x_l \) with reward \( r_l \). 
The loss function is defined as follows by minimizing the negative log-likelihood:

\vspace{-0.5cm}
\begin{equation}
\begin{split}
\mathcal{L}  &= -\log(P(x_h \succ x_l|c)) \\
&= - \bigg[\mathds{1}(x_1 \succ x_2) \log(P(x_1 \succ x_2|c)) \\
&\quad\;\; + \mathds{1}(x_2 \succ x_1) \log(P(x_2 \succ x_1|c) \bigg].
\end{split}
\end{equation}
\vspace{-0.5cm}

Uncertainty-aware ranking encourages the model to leverage the underlying distribution of pairwise data, instead of relying solely on a single scalar score. This approach enhances the model's ability to capture nuanced uncertainties in human annotations, ultimately improving the overall ranking accuracy.

\subsection{Enhancing Image Generation with CoHP}

We introduce CoHP, an approach to enhance image generation quality by iteratively refining outputs using \mname. 
As shown in \Cref{fig:cohp}, CoHP consists of the Model-wise Preference stage and Sample-wise Preference stage. 
In each stage, we employ \mname as the reward model to evaluate generated images and guide the iterative selection of the best candidates, improving generation quality over multiple rounds.

\input{tables/hpsv3_category}

\noindent\textbf{Model-wise Preference.} Given a specific prompt, different models may demonstrate varying levels of generation quality. Therefore, it is essential to first identify a model that best aligns with the given prompt. 
We denote the pool of candidate models as:
$
    \mathcal{M}=\{m_{i}\mid i=1,2,...,M\},
$
where $m_{i}$ indicates the $i$-th model in the pool, $M$ representing the total number of candidate models. We use each model $m_i$ to generate a set of candidate images across $N$ rounds with the given prompt:
$
    \mathcal{I} = \{I_{i,j} \mid i=1,2,...,M;, j=1,2,...,N\},
$
where $I_{i,j}$ denotes the image generated by model $m_{i}$ in round $j$. 
Each $I_{i,j}$ is then rated by HPSv3, yielding a preference score $r_{i,j}$. The average score for model $m_{i}$ is computed as:
$
\bar{r}_i = \frac{1}{N} \sum_{j=1}^{N} r_{i,j},
$
As the final step of the Model-wise selection, the model with the highest $\bar{r}_i$ is selected as the golden model $m^{*}$:
\begin{equation}
m^{*} = \arg\max_{i} \bar{r}_i.
\end{equation}

\noindent\textbf{Sample-wise Preference.} 
Given a prompt, the selected golden model $m^{*}$ generates a batch of $B$ images in each sampling round. We denote the batch of images generated in round $k$ as:
$
    I_k = \{I_{n,k} \mid n = 1, ..., B\},
$
where $k \in \{N+1, N+2, ..., N+S\}$ represents the current generation round in the Sample-wise stage, $n$ indexing the images in the batch, $S$ indicating the additional round numbers in the Sample-wise stage. Each image $I_{n,k}$ is evaluated by HPSv3, producing a set of scores $\{r_{n,k} \mid n = 1,...,B\}$.

The image with the highest HPSv3 score in round $k$ is selected as the reference image for the next generation round:
$
    I_k^\star = \arg\max_{n} r_{n,k}.
$
The selected image $I_k^\star$ will be blended with noise and combined with the original text prompt to form the input condition for the next generation round. This process repeats iteratively for additional $S$ rounds. Finally, the image with the highest overall score across all rounds is chosen as the final golden image:
\vspace{-0.2cm}
\begin{equation}
    I^\star = \arg\max_{n,k} r_{n,k}.
\end{equation}
\vspace{-0.3cm}

By leveraging HPSv3 for iterative refinement, CoHP ensures both the selection of the most suitable model and the generation of the highest-quality image. This structured approach enhances image quality and semantic alignment, offering a powerful alternative to conventional one-step generation methods.

%% file: tables/hpsv3_category.tex
\begin{table*}[!t]
\centering
\small

\scalebox{0.75}{
\setlength{\tabcolsep}{0.8mm}{
\begin{tabular}{l|cccccccccccccccccccccccccc}
\toprule
\textbf{Models} & \textbf{All} && \textbf{Characters} && \textbf{Arts} && \textbf{Design} && \textbf{Architechture} && \textbf{Animals} && \textbf{Natural Scenery} && \textbf{Transportation} && \textbf{Products} && \textbf{Plants} && \textbf{Food} && \textbf{Science} && \textbf{Others} \\
\midrule              
                               
    Kolors \cite{kolors}& \textbf{10.55} && \textbf{11.79} && \textbf{10.47} && \textbf{9.87} && \underline{10.82} && \textbf{10.60} && 9.89 && \underline{10.68} && \underline{10.93} && \textbf{10.50} && \textbf{10.63} && \underline{11.06} && \underline{9.51} \\
    Flux-dev~\cite{flux2024} & \underline{10.43} && \underline{11.70} && \underline{10.32} && 9.39 && \textbf{10.93} && \underline{10.38} && \underline{10.01} && \textbf{10.84} && \textbf{11.24} && \underline{10.21} && 10.38 && \textbf{11.24} && 9.16 \\
    Playground-v2.5~\cite{playground} & 10.27 && 11.07 && 9.84 && \underline{9.64} && 10.45 && \underline{10.38} && 9.94 && 10.51 && \underline{10.62} && 10.15 && \underline{10.62} && 10.84 && 9.39 \\
    Infinity~\cite{han2024infinity} & 10.26 && 11.17 && 9.95 && 9.43 && 10.36 && 9.27 && \textbf{10.11} && 10.36 && 10.59 && 10.08 && 10.30 && 10.59 && \textbf{9.62} \\
    CogView4~\cite{cogview3} & 9.61 && 10.72 && 9.86 && 9.33 && 9.88 && 9.16 && 9.45 && 9.69 && 9.86 && 9.45 && 9.49 && 10.16 && 8.97 \\
    PixArt-$\Sigma$~\cite{pixart} & 9.37 && 10.08 && 9.07 && 8.41 && 9.83 && 8.86 && 8.87 && 9.44 && 9.57 && 9.52 && 9.73 && 10.35 && 8.58 \\
    Gemini 2.0 Flash ~\footnotemark & 9.21 && 9.98 && 8.44 && 7.64 && 10.11 && 9.42 && 9.01 && 9.74 && 9.64 && 9.55 && 10.16 && 7.61 && 9.23 \\
    
    Stable Diffusion XL~\cite{podell2023sdxl} & 8.20 && 8.67 && 7.63 && 7.53 && 8.57 && 8.18 && 7.76 && 8.65 && 8.85 && 8.32 && 8.43 && 8.78 && 7.29 \\
    Hunyuan~\cite{hunyuandit} & 8.19 && 7.96 && 8.11 && 8.28 && 8.71 && 7.24 && 7.86 && 8.33 && 8.55 && 8.28 && 8.31 && 8.48 && 8.20 \\
    Stable Diffusion 3~\cite{sd3} & 5.31 && 6.70 && 5.98 && 5.15 && 5.25 && 4.09 && 5.24 && 4.25 && 5.71 && 5.84 && 6.01 && 5.71 && 4.58 \\

    Stable Diffusion v2.0~\cite{stable_diffusion} & -0.24 && -0.34 && -0.56 && -1.35 && -0.24 && -0.54 && -0.32 && 1.00 && 1.11 && -0.01 && -0.38 && -0.38 && -0.84 \\

\bottomrule
\end{tabular}
}}
\vspace{-2mm}
\caption{\label{tab:bench_mark_category}\textbf{HPDv3 Benchmark of popular image generation models.} We generate images for each participating model using its recommended inference settings or official API. The best and second-best results are highlighted in \textbf{bold} and \underline{underlined}, respectively. }
\vspace{-5mm}
\end{table*}

%% file: sec/experiment.tex
\section{Experiment}
\begin{figure}[!t]
  \centering
  \small
  \includegraphics[width=0.85\linewidth]{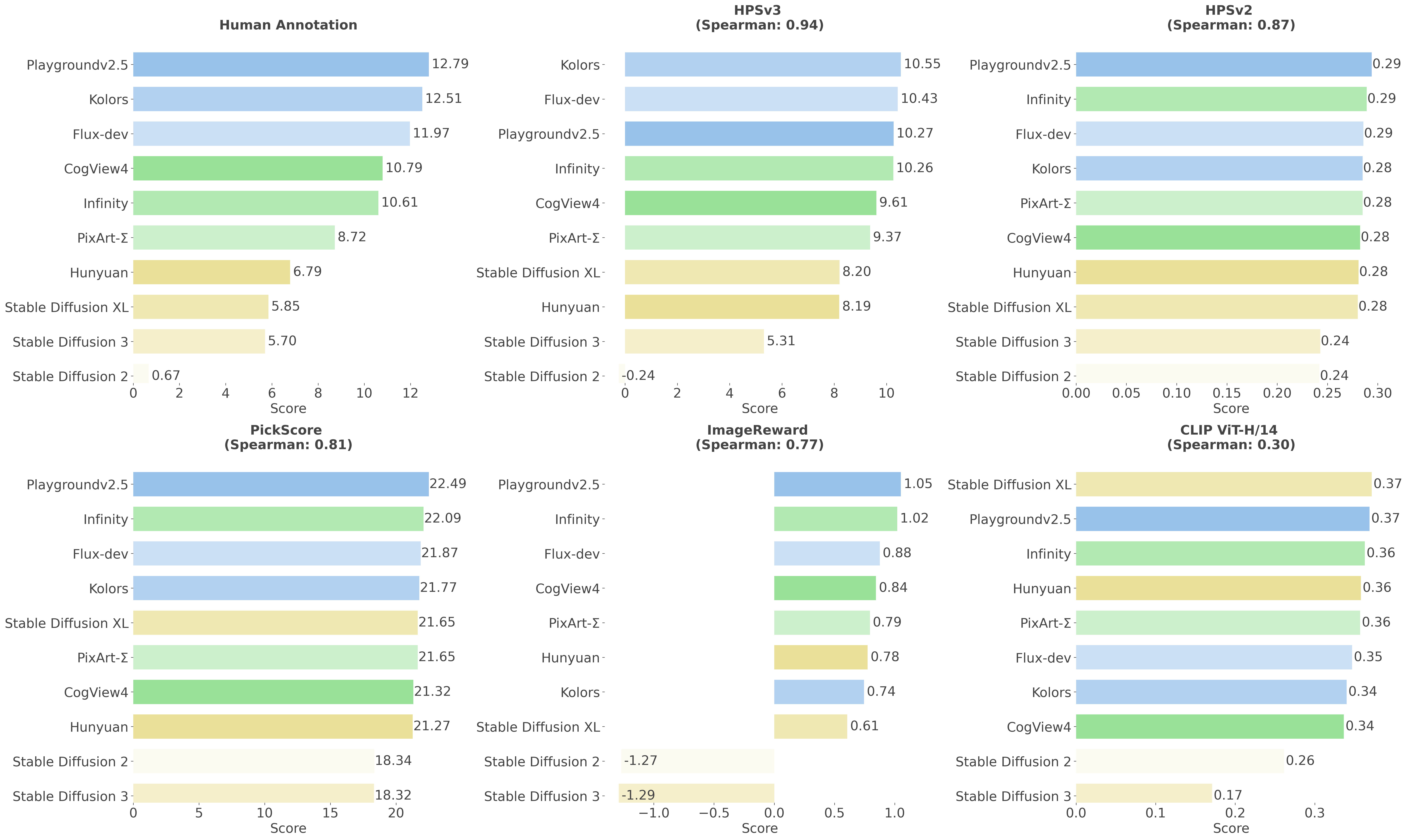}
  \caption{\textbf{Ranking of generative models across different metrics.} HPSv3 shows the highest correlation with human annotation, indicating its superior performance in reflecting human preferences.}
  \label{fig:model_metrics_comparison_shades}
  \vspace{-5mm}
\end{figure}
\footnotetext{https://deepmind.google/models/gemini/flash/}

\subsection{HPSv3}
\subsubsection{Implementation Details}
\label{implementation}
\noindent\textbf{HPDv3.} We generate images using various models, each configured with their recommended inference settings, running on 80GB NVIDIA A800 GPUs. For standard generation tasks, images are produced at each model’s recommended resolution, typically using square aspect ratios. When generating images from prompts derived from real images, we preserve the original aspect ratio to avoid annotation bias and to increase the diversity of resolutions in our training dataset. This strategy ensures a more varied and representative set of training samples that better reflect real-world image distributions.

\noindent\textbf{HPSv3.} We train \mname\ using $1.5$ million well-annotated pairwise samples. For HPSv3, we adopt Qwen2VL-7B as the backbone, with all parameters set as trainable. Training is performed over $2$ epochs using $48$ NVIDIA A800 GPUs (each with $80$GB of memory). The model is optimized with a learning rate of $2 \times 10^{-6}$, a warm-up ratio of $0.05$, and a total batch size of $384$ (corresponding to a per-GPU batch size of $8$). All training images are resized to $448 \times 448$ pixels while preserving their original aspect ratios. Additional training details are provided in the supplementary materials.

\input{tables/rank_analysis}
\subsubsection{HPDv3 Benchmark and Evaluation}

To establish a comprehensive benchmark for human preference, we sample $1,000$ prompts from each category of HPDv3, resulting in a total of $12,000$ prompt entries. Using $11$ widely adopted generative models, we generate images based on these prompts, producing $132,000$ text-image pairs, which constitute the \dname\ Benchmark. \Cref{tab:bench_mark_category} reports the \mname\ scores on this benchmark. Since \mname\ is trained on both generated and high-quality real images, the resulting model rankings align with expectations, demonstrating \mname's robustness and its effectiveness as a comprehensive metric for evaluating current and emerging image generation models.

\subsubsection{Preference Comparison}

\noindent\textbf{Comparison between HPSv3 and other evaluation methods on HPDv3 benchmark.} 
We evaluate various preference models across a range of generative models, as shown in~\Cref{fig:model_metrics_comparison_shades} and~\Cref{tab:metrics_comparison_rank}. HPSv3 consistently demonstrates strong alignment with human preferences, achieving the highest correlations (Spearman \( r = 0.94 \), Kendall \( \tau = 0.8222 \)) and effectively distinguishing between models across the performance spectrum. HPSv2 also shows reasonable consistency but with limited discriminative power (\eg, identical scores for Playgroundv2.5, Infinity, and Flux-dev). PickScore performs well on low-quality models but struggles with mid-tier cases, while ImageReward, despite capturing variation, deviates from human rankings. CLIP exhibits the weakest correlation, underscoring its limited ability to reflect human preference. Overall, HPSv3 emerges as the most reliable metric for evaluating generative models.

\input{tables/vsbenchmark}

\noindent\textbf{Performance on various datasets compared with other models.} 
As presented in \Cref{tab:vsbenchmark11}, \mname demonstrates outstanding performance, achieving state-of-the-art accuracy scores of $72.8\%$, $85.4\%$, and $76.9\%$ on the PickScore, HPDv2, and HPDv3 testsets, respectively.
Notably, on the challenging HPDv3 dataset, the accuracy of \mname shows a growth margin of $19.3\%$  over HPSv2 and $17.2\%$ over PickScore.
Moreover, PickScore suffers a sharp drop of $14.2$ percent point (pp) drop, and MPS declines by $19.1$pp when moving from HPDv2 to HPDv3, whereas \mname remain remarkably stable with only a slight decrease of $8.5$pp.
These results underscore HPSv3’s superior ability to model human preferences and its robustness across datasets, while also demonstrating that HPDv3 poses substantial challenges for existing methods.
\subsubsection{Ablation Study}

We conduct a comparative analysis using different backbone architectures ( Qwen2VL-7B, Qwen2VL-2B and CLIP) under identical training conditions as detailed in Sec. \ref{implementation}. As shwon in \Cref{tab:combined_ablation}, adopting Qwen2VL-7B as the backbone yields substantial performance improvements, with notable gains of $10.6$ percent point (pp) and $13.4$ pp on the HPDv3 over Qwen2VL-2B and CLIP, respectively. Notably, with the uncertainty-aware ranking loss, the accuracy improves across all datasets. Notably, the accuracy on the PickScore test set demonstrates a significant $2.88\%$ increase, highlighting the effectiveness and robust performance of the uncertainty-aware ranking loss. 

\begin{figure}[t]
  \centering
  \small
  \includegraphics[width=\linewidth]{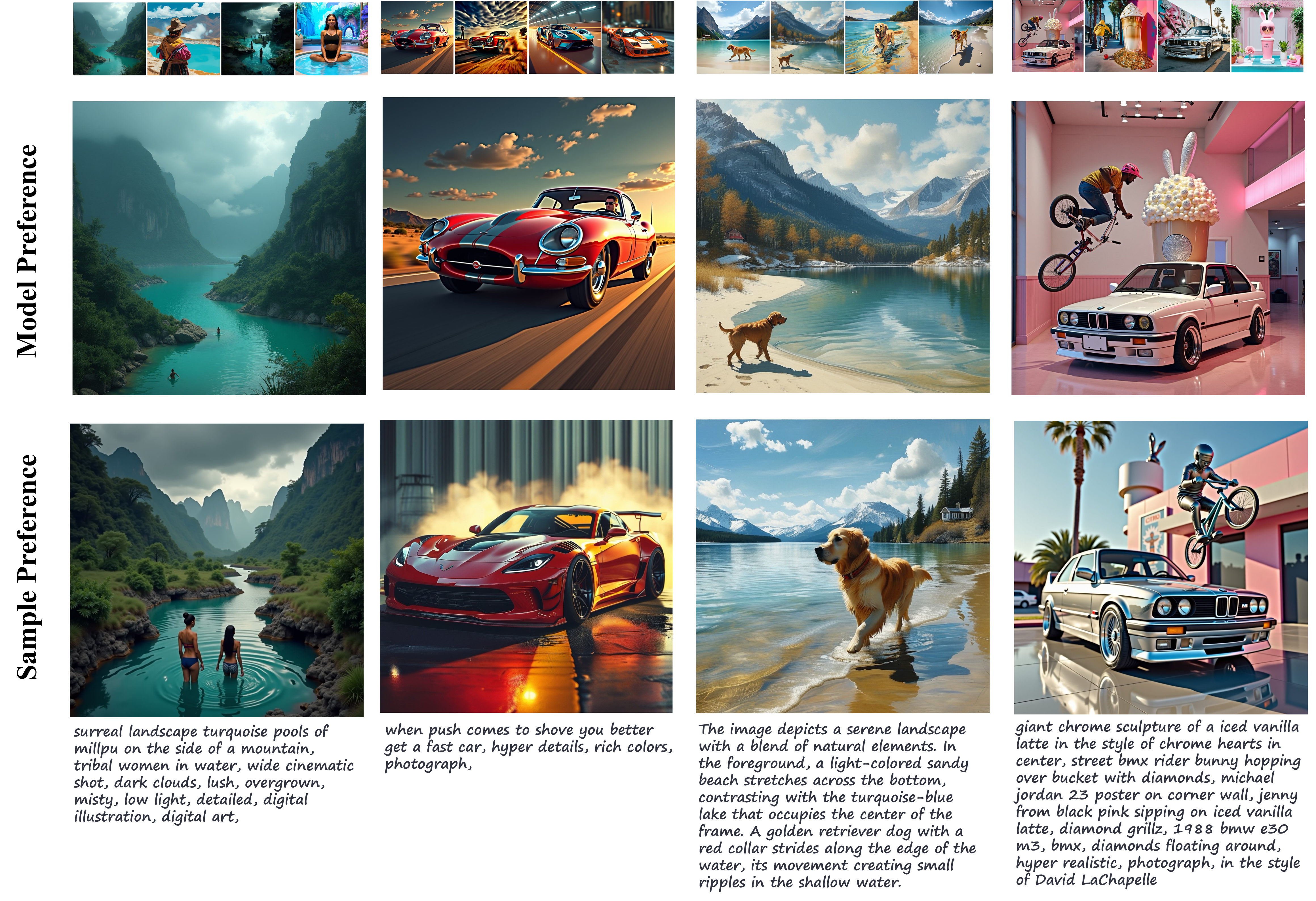}
  \vspace{-8mm}
  \caption{\textbf{Image generation with Chain-of-Human-Preference (CoHP).} It illustrates that the quality of the generated images improves progressively through CoHP. The first row displays all the candicate images generated by participant models (Flux-dev, Kolors, Playground v2.5, and SD3).}
  \label{fig:cohp_result}
\end{figure}

\input{tables/backboneablation}
\subsection{CoT with HPSv3}
\subsubsection{Inference Settings}
In this work, we adopt four generative models, Flux-dev~\cite{flux2024}, Playground v2.5~\cite{playground}, SD3~\cite{sd3}, and Kolors~\cite{kolors}, as candidate models within our Chain-of-Thought Human Preference (CoHP) framework. The CoT process is configured with four sequential steps for both the Model-wise and Sample-wise stages. High-resolution images are generated at $1024 \times 1024$ pixels using each model’s recommended settings. In the Model-wise stage, a batch of four images is generated per model to support the four-round evaluation. In the Sample-wise stage, the denoising strength is set to $0.8$ for the first two rounds and $0.5$ for the remaining rounds.

\subsubsection{Image Generation with CoHP}

\Cref{fig:cohp_result} illustrates the progression of generated images through CoHP-HPSv3. During the Model-wise stage, the first row displays the best images generated by each model for comparison, demonstrating that the chosen model achieves superior performance in both image quality and text-image alignment. In the Sample-wise stage, HPSv3 functions as a reliable metric to refine images further, enhancing their details and structural coherence.

\begin{figure}[t]
  \centering
  \small
  \includegraphics[width=1.0\linewidth]{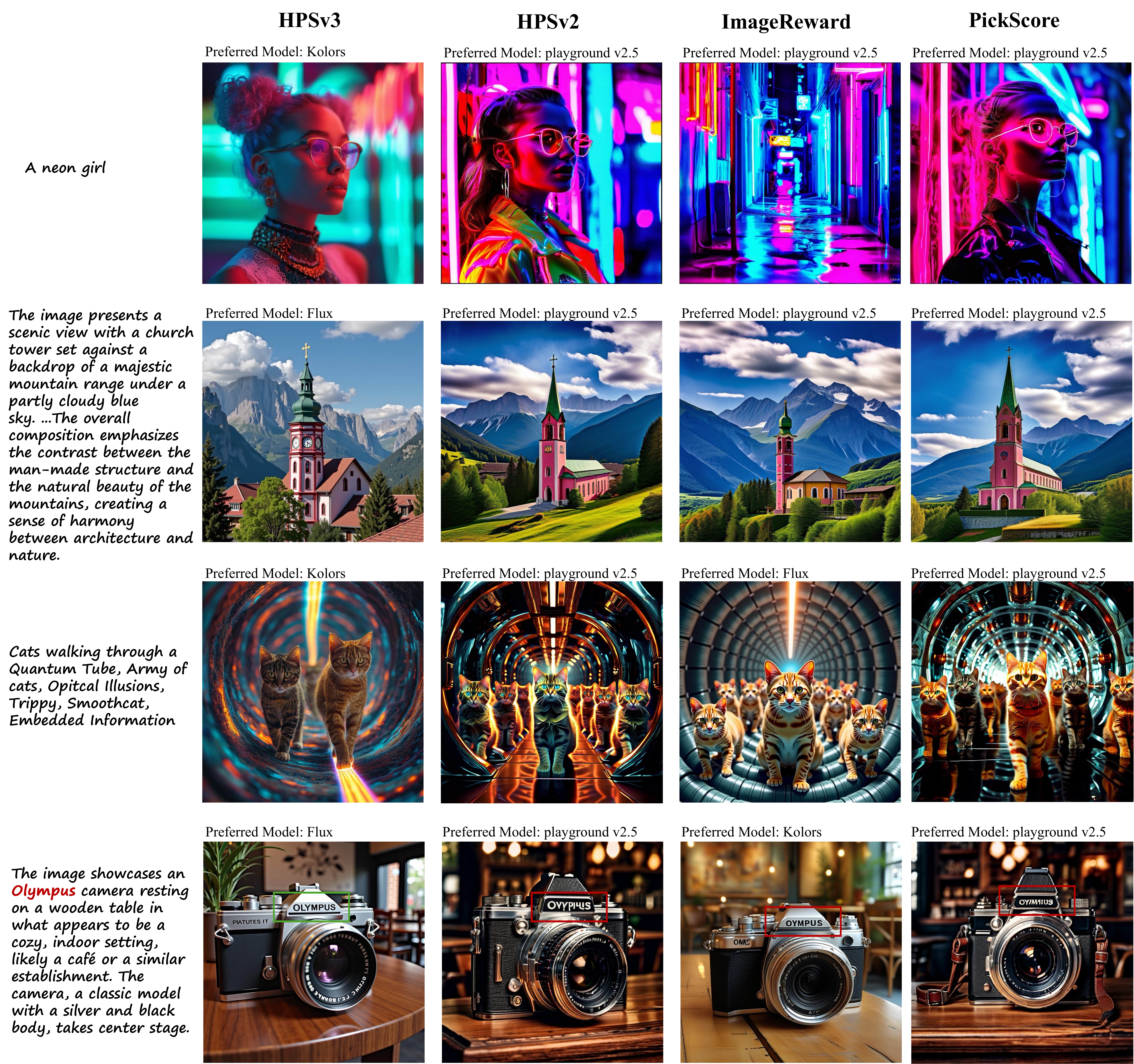}
  \caption{\textbf{Comparison of CoHP results with different preference models.} CoHP-HPSv3 shows the best coherence and fidelity, while others suffer from oversaturation or instability.}
  \label{fig:cohp_compare}
  \vspace{-0.6cm}
\end{figure}

\subsubsection{Qualitative evaluation}
We experiment with different human preference models in CoHP to explore their performance. CoHP-\mname is capable of selecting the model that delivers the best performance in the Model-wise stage and generating highly detailed images in the Sample-wise stage. As shown in \Cref{fig:cohp_compare}, CoHP-PickScore generates images with excessive color saturation and semantic inconsistencies. CoHP-ImageReward and CoHP-HPSv2 produce some impressive results but suffer from generation instability. In contrast, CoHP-\mname produces more faithful images with superior coherence and better text-image alignment compared to other methods.
\subsubsection{User Study}
\begin{figure}[t]
  \centering
  \small
  \includegraphics[width=1.0\linewidth]{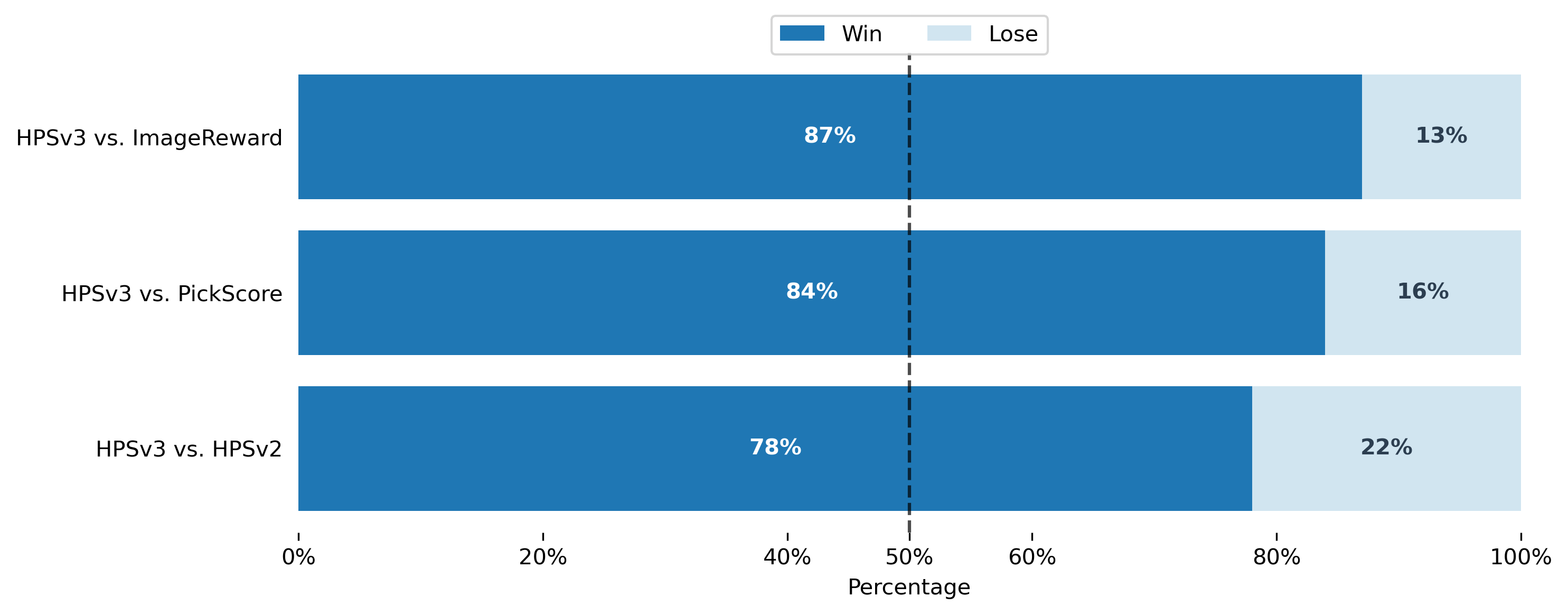}
  \caption{\textbf{Human evaluation on CoHP with different human preference models.} HPSv3 outperforms all other human preference models by a large margin.}
  \vspace{-2mm}
  \label{fig:human_evalution}
\end{figure}

We conduct a human evaluation on $100$ generated text-image pairs to compare HPSv3 against HPSv2, ImageReward, and PickScore. We perform pairwise comparisons between our CoHP-HPSv3 approach and each baseline (CoHP-HPSv2, CoHP-ImageReward, and CoHP-PickScore) to assess relative performance gains. As shown in \Cref{fig:human_evalution}, HPSv3 outperforms Imagereward with a $87\%$ win rate and demonstrates a clear advantage over PickScore and HPSv2. These results confirm that HPSv3 more effectively captures human preferences than existing models.

\subsubsection{Ablation Study}

\input{tables/round}
We randomly sample $1,000$ prompts from the HPDv3 Benchmark and generate images across different iteration rounds for each stage. As shown in \Cref{tab:round}, for the Model-wise stage, the HPSv3 score improves from $11.34$ in Round $1$ to $11.69$ in Round $4$, indicating that additional rounds contribute to better model selection initially. However, the score remains stable after Round $4$, with only a slight decrease to $11.65$ in Round $5$, suggesting diminishing returns with more rounds.

For the Sample-wise stage, the HPSv3 score consistently improves from Round $1$ to Round $4$, with only a marginal drop in Round $5$. This trend demonstrates that increasing the number of rounds in this stage enhances sample quality up to a certain point. Based on this analysis, we conclude that $4$ rounds strike a balance between computational cost and performance, which we adopt as the setting in our paper.

%% file: tables/rank_analysis.tex
\begin{table}[t]
\centering
\small
\scalebox{0.8}{
\begin{tabular}{lccc}
\toprule
\textbf{Metric} & \textbf{Spearman ($r$)}$\uparrow$  & \textbf{Kendall ($\tau$)}$\uparrow$   & \textbf{Normalized MSE}$\downarrow$  \\
\midrule
CLIP          & 0.30   & 0.16  & 0.100  \\
Aesthetic     & 0.76   & 0.60  & \underline{0.036}  \\
ImageReward   & 0.77  & 0.64   & 0.056 \\
PickScore     & 0.81   & 0.63   & 0.042 \\
HPSv2         & \underline{0.87}   & \underline{0.76} & 0.038   \\
HPSv3(Ours)   & \textbf{0.94}  & \textbf{0.82} & \textbf{0.029} \\
\bottomrule
\end{tabular}}
\vspace{-2mm}
\caption{\textbf{Correlation between automated metrics and human preference rankings.} HPSv3 shows the highest alignment with human judgments, followed by HPSv2 and PickScore. In contrast, CLIP exhibits the weakest correlation.}
\label{tab:metrics_comparison_rank}
\vspace{-5mm}
\end{table}

%% file: tables/vsbenchmark.tex
\begin{table}

  \centering
  \setlength\tabcolsep{0.9mm}{
  \scalebox{0.76}{
  \begin{tabular}{lccccc}
    \toprule
    \textbf{Model} & \textbf{ImageReward} & \textbf{PickScore} & \textbf{HPDv2} & \textbf{HPDv3} \\
    \midrule
    CLIP ViT-H/14~\cite{clip} & 57.1 & 60.8 & 65.1 & 48.6\\
    Aesthetic Score Predictor~\cite{schuhmann2022laion} & 57.4 & 56.8 & 76.8 & 59.9\\
    ImageReward~\cite{xu2023imagereward} & 65.1 & 61.1 & 74.0 & 58.6\\
    PickScore~\cite{kirstain2023pick} & 61.6 & \underline{70.5} & 79.8 & \underline{65.6}\\  
    HPS~\cite{wu2023better} & 61.2 & 66.7 & 77.6 & 63.8\\
    HPSv2~\cite{hpsv2} & 65.7 & 63.8 & 83.3 & 65.3\\
    MPS~\cite{mps} & \textbf{67.5} & 63.1 & \underline{83.5} & 64.3\\
    \textbf{HPSv3 (Ours)} & \underline{66.8} &  \textbf{72.8} &  \textbf{85.4} &  \textbf{76.9}\\
    \bottomrule
  \end{tabular}
  }}
  \caption{\label{tab:vsbenchmark11}\textbf{Preference prediction accuracy (\%) on the test sets of ImageReward, HPDv2 and HPDv3.} The best and second-best results are \textbf{bolded} and \underline{underlined}. HPSv3 exhibit exceptional confidence in human preference. }
  \vspace{-5mm}
\end{table}

%% file: tables/backboneablation.tex
\begin{table}[t!]
  \centering
  \setlength\tabcolsep{0.8mm}{
  \scalebox{0.76}{
  \begin{tabular}{lccccc}
    \toprule
    \textbf{Backbone} & \textbf{Loss}& \textbf{IR} & \textbf{PickScore} & \textbf{HPDv2} & \textbf{HPDv3} \\
    \midrule
    CLIP ViT-H/14~\cite{clip}  & - & 62.4 & 63.9 & 82.3 & 63.5\\
    QWen2VL-2B~\cite{qwen2vl}  & uncertainty & 57.9 & 63.6 & 80.8 & 66.3\\
    QWen2VL-7B~\cite{qwen2vl}  & ranknet loss & 66.1 & 70.6 & 85.3 & 76.3\\
    QWen2VL-7B~\cite{qwen2vl}  & uncertainty & \textbf{66.8} & \textbf{72.8} & \textbf{85.4} & \textbf{76.9}\\
    \bottomrule
  \end{tabular}
  }}
\caption{\textbf{Ablation study on model backbones and loss function.} Results are reported as preference accuracy (\%) on ImageReward(IR), Pickapic and HPDv2 and HPDv3 testsets. The best results are highlighted in bold.}
\label{tab:combined_ablation}
\vspace{-5mm}
\end{table}

%% file: tables/round.tex
\begin{table}[!t]
  \centering
  \small
  \setlength\tabcolsep{1.2mm}{
  \scalebox{0.8}{
  \begin{tabular}{lccccc}
    \toprule
     & \textbf{Round 1} & \textbf{Round 2} & \textbf{Round 3} & \textbf{Round 4} & \textbf{Round 5} \\
    \midrule
    \textbf{Model-wise Preference}& 11.34 & 11.46 & 11.68 & \textbf{11.69} & 11.65\\
    \midrule
    \textbf{Sample-wise Preference}& 11.59 & 12.69 & 12.64 & \textbf{12.84} & 12.82\\
    \bottomrule
  \end{tabular}
  }}
\caption{\textbf{Ablation study on the round number.} We calculate the HPSv3 score with different rounds of the preference stages, opting to run each stage for $4$ rounds.}
\label{tab:round}
\vspace{-5mm}
\end{table}

%% file: sec/conclusion.tex
\section{Conclusion}
We introduce Human Preference Score v3 (\mname), involving: (1) \dname, the first wide-spectrum human preference dataset with $1.08$ millon text-image pairs and $1.17$ millon annotated comparisons from state-of-the-art generative models and high-quality real images, and (2) a preference model leveraging VLM-based feature extraction with uncertainty-aware ranking loss for accurate image scoring. Additionally, we propose Chain-of-Human-Preference (CoHP), a novel reasoning-based approach for iterative image refinement without requiring extra training data. Extensive experiments validate \mname as a robust benchmark and CoHP as an efficient, human-aligned method for enhancing image generation quality.

%% file: sec/acknowledgement.tex
\section*{Acknowledgement}
We express our gratitude to Xuan Ouyang and Wenting Xu for their helpful discussion.
We appreciate Changyao Tian, who assisted in the data migration of this project.
We gratefully acknowledge Fuxi Youling Crowdsourcing for mobilizing a substantial team of annotators and reviewers to carry out the extensive annotation efforts required for this work.
This project is funded in part by National Key R\&D Program of China Project 2022ZD0161100, by the Centre for Perceptual and Interactive Intelligence (CPII) Ltd under the Innovation and Technology Commission (ITC)’s InnoHK, and in part by Guangdong Basic and Applied Basic Research Foundation (No. 2023B1515130008, XW).

%% file: sec/X_suppl.tex
\section{Image Sources of HPDv3}
\input{tables/sourcemodel}

\begin{figure}[!t]
  \centering
  \small
  \includegraphics[width=\linewidth]{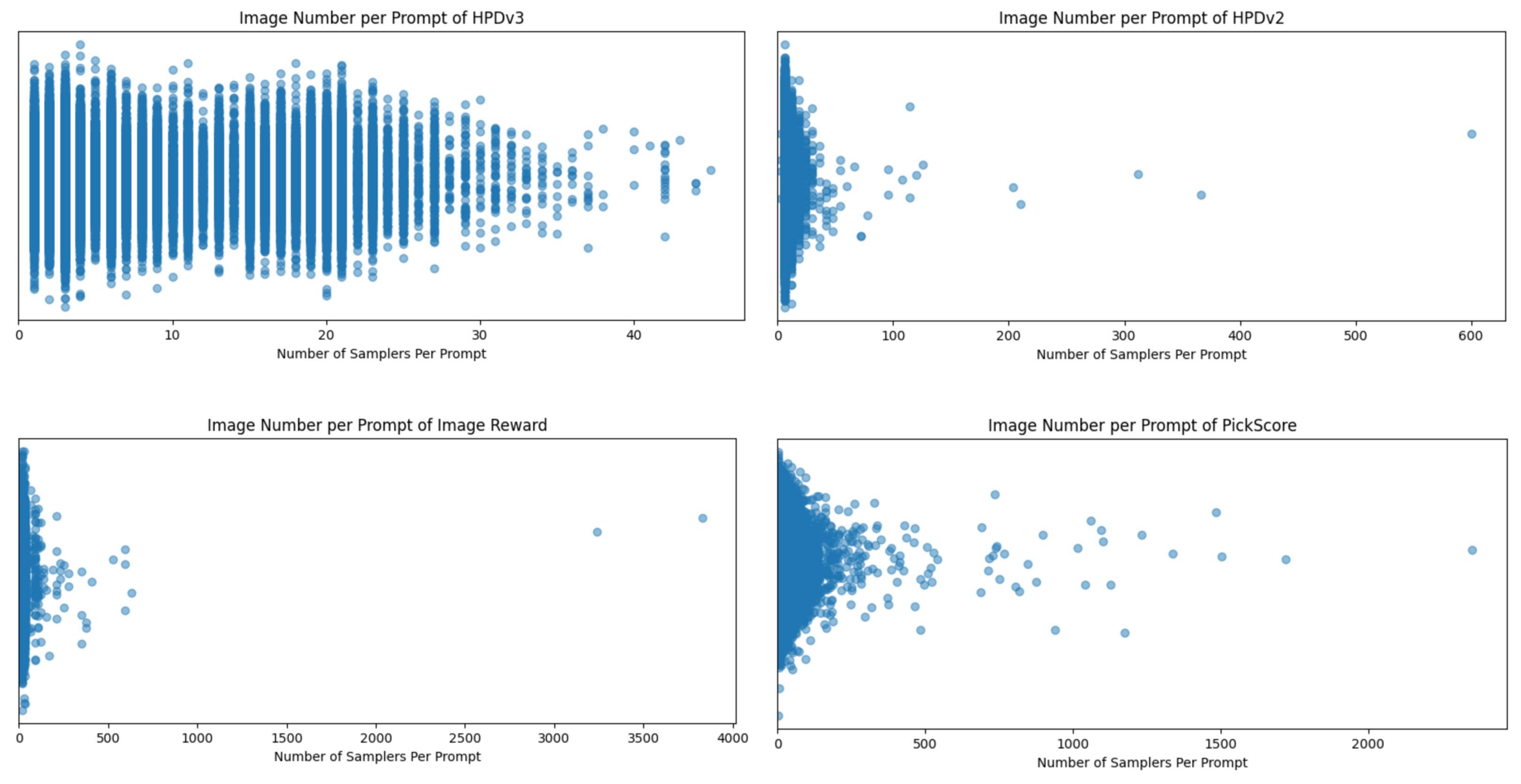}
  \caption{\textbf{Image numbers per prompt of each dataset.}}
  \label{fig:prompt_distribution}
\end{figure}

\Cref{tab:sourcemodel} summarizes the source models and images in HPDv3. Our dataset includes outputs from recent state-of-the-art image generation models, high-quality real-world images, and images generated by Midjourney, resulting in a total of $1.08$M text-image pairs. Additionally, we compare the text-image pairs in HPDv3, HPDv2, PickScore, and ImageReward datasets. \Cref{fig:prompt_distribution} reveals that HPDv2~\cite{hpsv2}, PickScore~\cite{kirstain2023pick}, and ImageReward~\cite{xu2024imagereward} datasets often associate identical prompts with more than $100$ images, leading to an uneven distribution with significant outliers. Such imbalances can negatively affect model training. In contrast, HPDv3 maintains a more balanced distribution, with no prompt linked to more than $50$ images, ensuring consistent and unbiased training for learning user preferences.

\section{Category distribution of HPDv3}
To better reflect user preferences for prompt categories, we categorize user prompts in JourneyDB~\cite{journeydb} into $12$ distinct classes, ensuring that the class proportions in HPDv3 closely match those in JourneyDB.

\begin{figure}[!t]
  \centering
  \small
  \includegraphics[width=\linewidth]{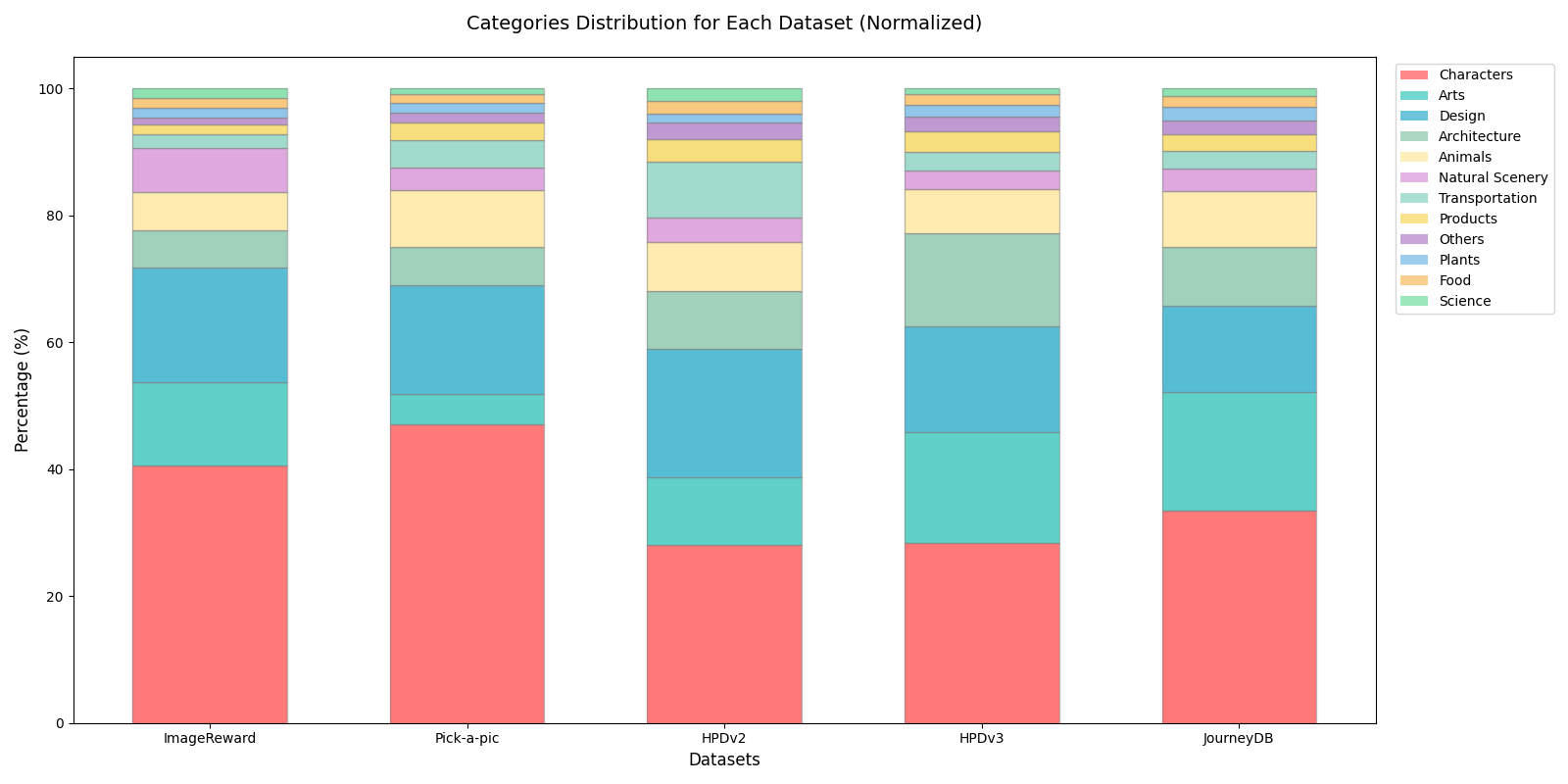}
  \caption{\textbf{Distribution of real images across $12$ categories collected from the Internet.}}
  \label{fig:supp_cate}
\end{figure}

As shown in \Cref{fig:supp_cate}, we compare the category distributions of HPDv3, HPDv2, ImageReward, and Pick-a-Pic datasets. The result shows that HPDv3 aligns most closely with the category proportions of JourneyDB, indicating that HPDv3 effectively captures user preferences for prompt categories, making it more representative and balanced.
\section{\dname Dataset Construction}

\subsection{Real Image Collection}
We collect aound $5$M high-quality real images from the Internet, covering a wide range of categories such as architecture, people, objects, animals, plants, landscapes, products, and posters. This diverse sampling ensures that the dataset is broad and representative. 

The collected images are predominantly authentic photographs. This forms a strong foundation for a high-quality dataset. However, despite our careful collection process, the dataset still contains some noise and irrelevant samples that require further refinement. 
\subsection{Aesthetic Model Training}
To efficiently filter large volumes of images based on their aesthetic quality, we develop a specialized visual assessment model. As our analysis reveals, the model trained by open-source community~\cite{improvedaestheticpredictor} exhibits a strong preference for oil painting, which may not accurately reflect human aesthetic preferences. This bias could potentially skew our image quality assessments toward a particular visual style rather than capturing more universal aspects of image quality. To address this limitation and better align with diverse human aesthetic judgments, we decide to retrain the model using our newly developed dataset. 

\noindent\textbf{Model Training}
We keep the model architecture similar to the open-source aesthetic model~\cite{improvedaestheticpredictor} but refine it to address aesthetic bias. We train our model on a single NVIDIA A100 80GB GPU using our carefully curated $20,000$-image annotation dataset. The training configuration employs a learning rate of $5\times10^{-3}$ and a batch size of $256$.

\subsection{High-quality Real-image Selection}
Using our trained aesthetic model, we evaluate the quality of all collected images. Fisrt, we filter out all images with a quality score below $4.0$, as these are consistently lower quality. Then, to ensure category diversity, we apply category-specific evaluation and proportional selection, focusing on the highest-scoring images within each category while maintaining the ratios of each category in HPDv3.

\begin{figure}[!t]
  \centering
  \small
  \includegraphics[width=0.8\linewidth]{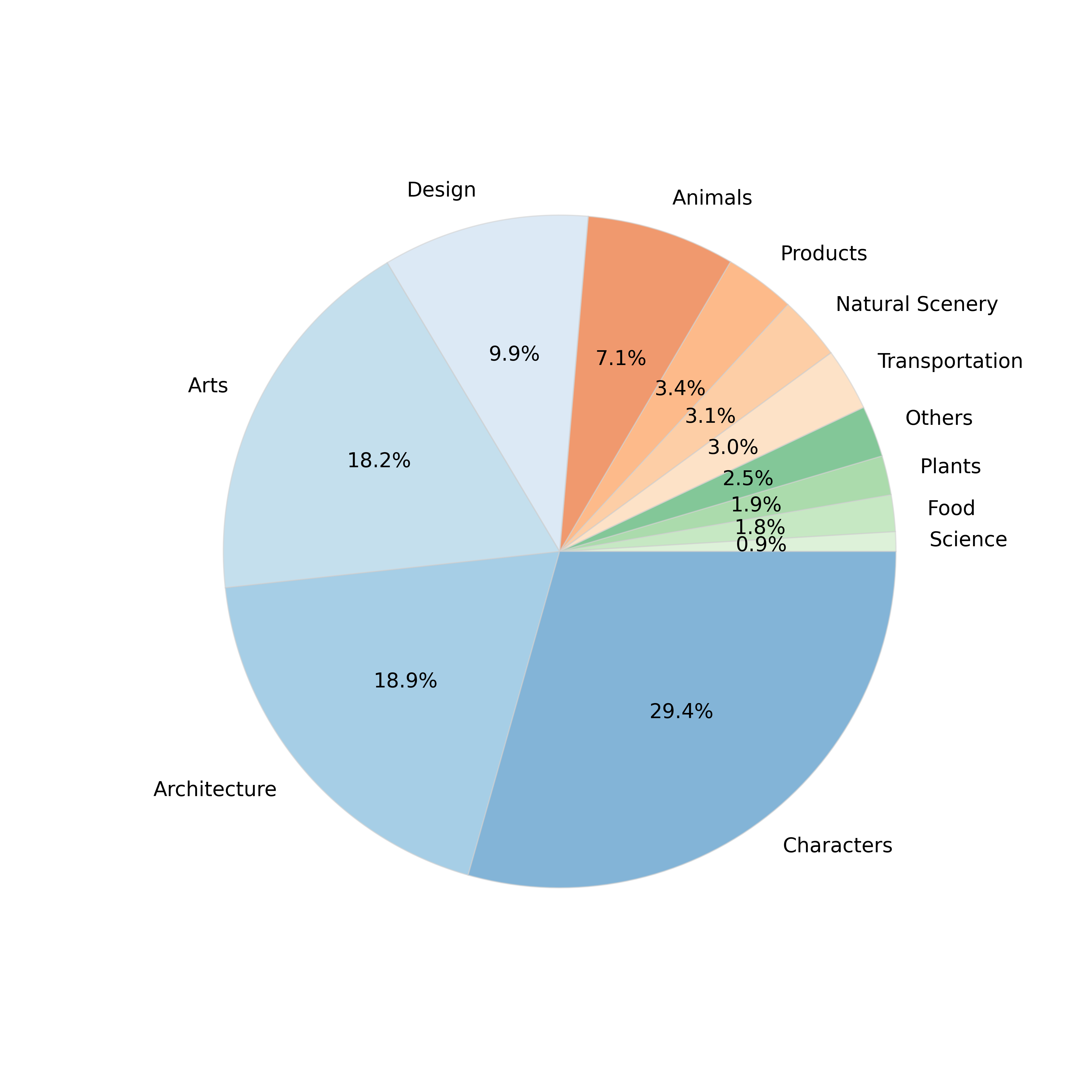}
  \caption{\textbf{Category distribution of high-quality real images in HPDv3 dataset.}}
  \label{fig:realimage_category}
\end{figure}

This process results in a final curated dataset comprising $58$k high-quality real images. The category distribution of these selected images, shown in \Cref{fig:realimage_category}, closely aligns with the proportions seen in \Cref{fig:supp_cate}, ensuring a well-balanced and high-quality dataset.

\subsection{Pairwise Image Generation}
We use various generative models with their recommended configurations to create images based on prompts from HPDv2, descriptions of real images, or JourneyDB. For images generated from HPDv2 and JourneyDB prompts, we use square dimensions to maintain consistency. For images based on real image descriptions, we match the aspect ratios of the original images to preserve their structure and visual integrity.
assessment of content and aesthetic qualities independent of image proportions.

After generation, we group images by the same prompt into pairs comparing outputs from different models. These pairwise comparisons allow us to evaluate the relative performance of generative models under identical prompts, providing more detailed insights into their strengths and weaknesses. This pairwise approach forms the foundation for further annotations and model training.
\input{tables/supp_score_standard}
\begin{figure}[!t]
  \centering
  \small
  \includegraphics[width=1\linewidth]{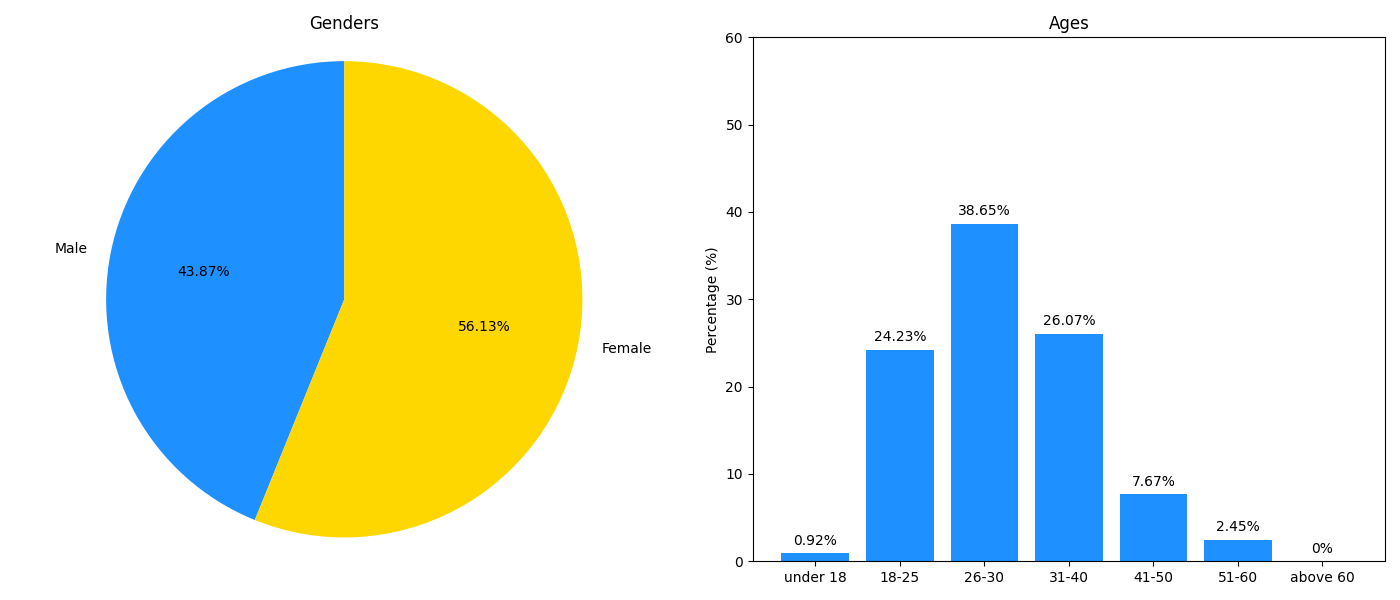}
  \caption{\textbf{Demographic Profile of Annotators: Gender Distribution and Age Stratification.}}
  \label{fig:gender_age}
\end{figure}
\begin{figure}[!t]
  \centering
  \small
  \includegraphics[width=1\linewidth]{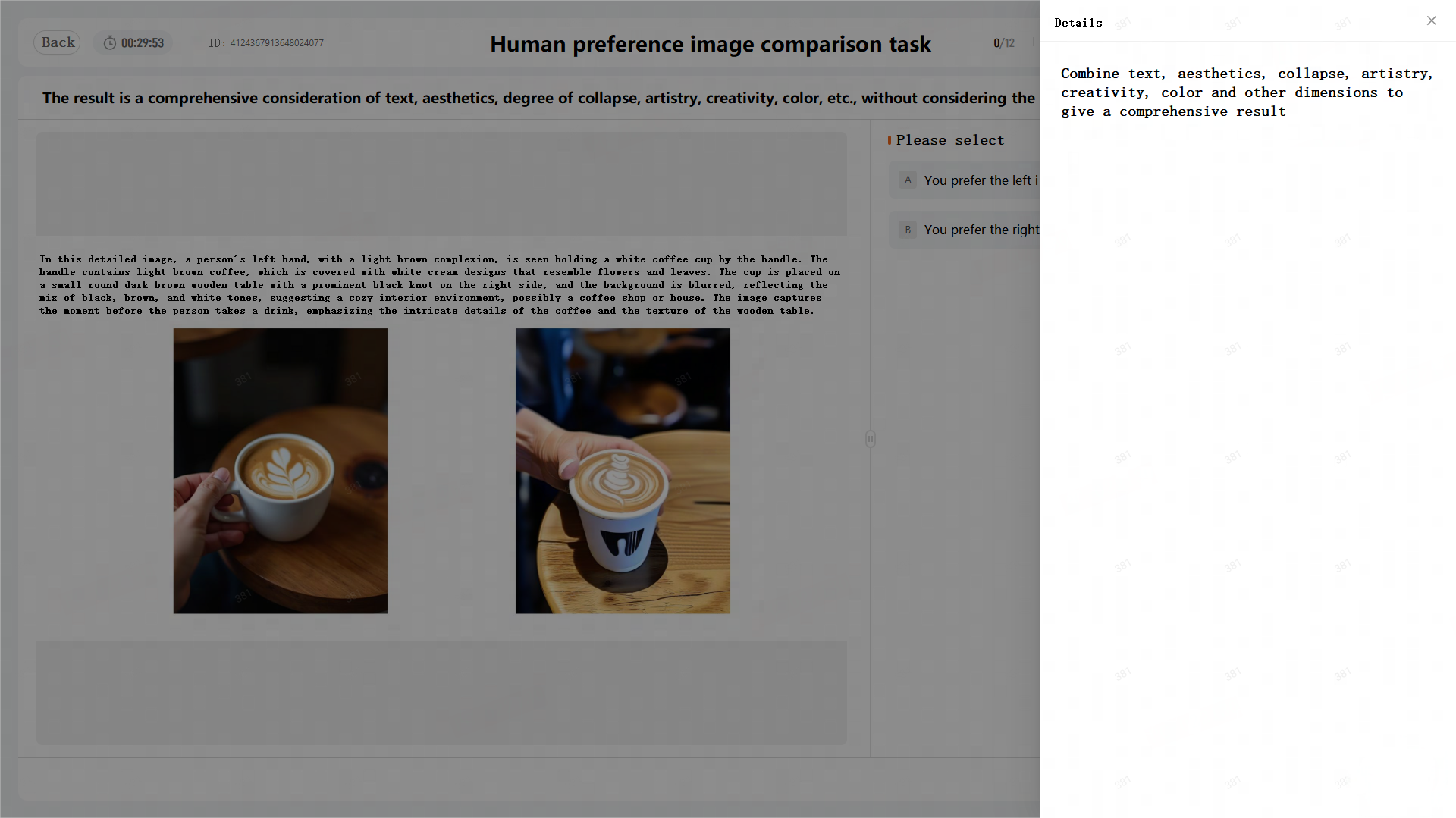}
  \caption{\textbf{Annotation interface of pairwise image comparison.}}
  \label{fig:annotation_interface}
\end{figure}
\section{Annotation Details}
\subsection{Image Annotation.}
To build a reliable training dataset, we subject the pairwise image data to thorough human annotation, following a standardized evaluation protocol.

As shown in \Cref{tab:score}, human evaluators are provided with detailed guidelines that define clear criteria for judgment. This structured approach helps ensure consistent annotations while capturing the multidimensional aspects of human preferences.

Each image is scored by $9-19$ experts, with an inter-annotator agreement threshold set at $0.9$ to ensure high reliability. The annotation process follows the same rigorous methodology described in the pairwise data annotation section of the main paper. By maintaining consistent evaluation standards across all stages of dataset creation, we ensure the overall quality and reliability of the annotated data.

\input{tables/supp_annotate_result}

\subsection{Demographic of annotators.}As shown in \Cref{fig:gender_age}, our annotation team has a relatively balanced gender distribution, with $56.13\%$ female and $43.87\% $male participants. The age demographics show that most annotators fall within the young to middle-aged categories, with $88.95\%$ aged between $18$ and $40$ years, and the $21-30$ age group being the largest ($38.65\%$).
This demographic composition offers several advantages, such as a strong familiarity with modern language patterns and current fashion trends. Additionally, the annotators come from diverse professional backgrounds, including college students, freelancers, artists, teachers, and engineers. This diversity brings a wide range of perspectives to the annotation process.
Such a well-balanced and diverse group ensures that the annotations capture the preferences of fashion-conscious consumers effectively, enhancing both the quality and relevance of our annotated dataset.

\subsection{Annotation interface and guidelines.}Figure \ref{fig:annotation_interface} illustrates our user-friendly annotation interface for pairwise image comparison. The interface displays two images generated from the same textual prompt, along with the prompt itself to provide contextual clarity. Annotators are required to select the image they prefer based on clearly defined evaluation guidelines.

The annotation protocol guides evaluators to assess images across three key dimensions: 
\begin{itemize}
    \item \textbf{Prompt Alignment}: How well the image matches the given textual description.
    \item \textbf{Aesthetic Quality}: The visual appeal and technical execution of the image.
    \item \textbf{Overall Coherence}: The logical consistency and naturalness of the scene depicted.

\end{itemize}
These structured criteria ensure that the annotations capture meaningful differences in quality while minimizing subjective bias. \Cref{tab:annotation_result} provides examples of annotation outcomes, including final preferences, confidence scores, and an assessment of annotator quality. This systematic approach helps maintain the reliability and consistency of the dataset annotations.

\subsection{Convergency of Annotations.}

\begin{figure}[!t]
  \centering
  \small
  \includegraphics[width=1\linewidth]{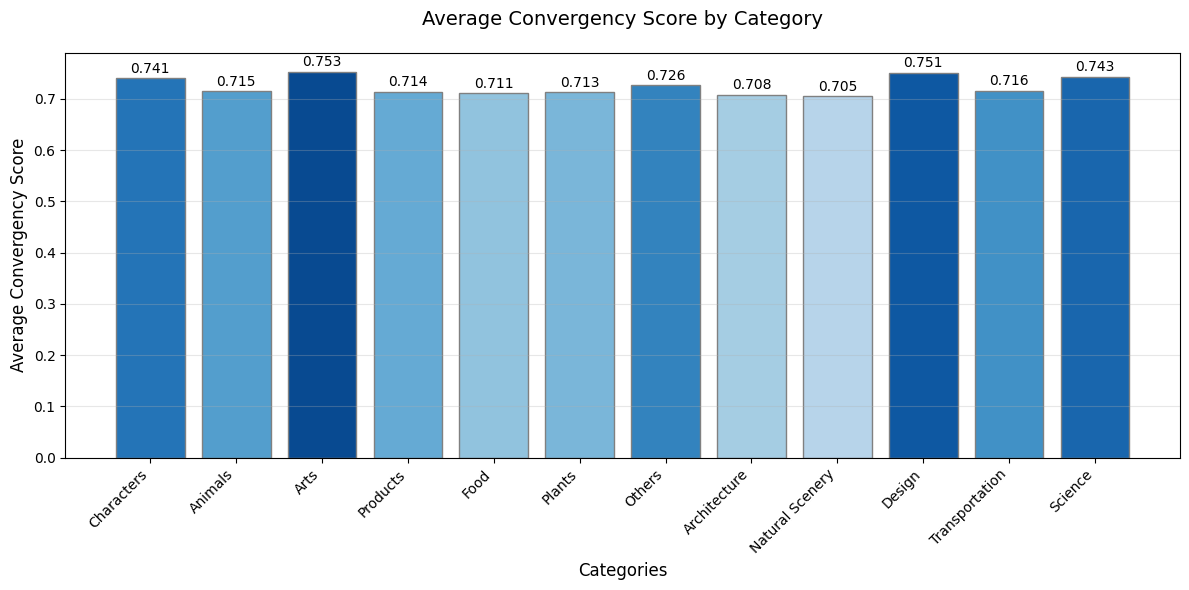}
  \caption{\textbf{Average convergency score by categories.}}
  \label{fig:convergency_by_category}
\end{figure}

We evaluate the annotation convergence for each data category to measure the consistency of annotator decisions. \Cref{fig:convergency_by_category} visualizes the level of convergence of annotations in all categories. Convergence is calculated by assessing the agreement among annotators when evaluating image pairs with the same text prompt. 

\section{\mname Training Details}
\input{tables/train_data}

\subsection{Training Dataset}
For training our final model, we use data from four sources: HPDv3, subsets of Pick-A-Pic and ImageReward, and real user preference data collected from Midjourney. In total, the training dataset comprises $1.5$ million pairwise samples.

Specifically, the HPDv3 dataset is composed of two parts. The first part includes pairwise comparisons of real images and images generated based on prompts from JourneyDB and HPDv2. These comparisons are annotated using the annotation pipeline, and only those with a confidence score of $0.95$ or higher are selected for training. 

The second part is a manually curated golden training set. This contains high-quality sample data. We use this golden trainset to train a filtering model, which shares the same architecture and training methodology as HPSv3. This filtering model is applied to choose samples from HPDv2. And we randomly select $250,000$ pairs from samples picked from HPDv2. This process enriches the HPDv3 dataset by increasing both model diversity and pairwise data diversity. To further boost the contribution of the golden set, we duplicate some of its samples during the training process.

Additionally, we include $350,000$ pairwise samples from Pick-A-Pic and $120,000$ samples from ImageReward. These datasets provide additional variety to enhance model performance.

Furthermore, we collect $150,000$ pairs of real user-choice data from Midjourney via the Internet. This real-world preference data is crucial for improving HPSv3’s ability to handle user selections during CoHP.

In summary, as detailed in \Cref{tab:training_data}, the training dataset is a diverse and comprehensive mix of high-quality, curated, and real-world user preference data. This careful composition ensures robust and adaptable model performance.

\subsection{Training on other datasets}
\input{tables/datasetablation}
\Cref{tab:datasetablation} presents the results of the dataset ablation study, where HPSv3 is trained on different datasets, including HPDv2, ImageReward, PickScore, and HPDv3. The evaluation is conducted across four metrics: ImageReward, PickScore, HPDv2, and HPDv3. Among the datasets, HPSv3 trained with HPDv3 outperforms others, achieving the best performance across all test sets—$66.8\%$ on ImageReward, $72.8\%$ on PickScore, $85.4\%$ on HPDv2, and $76.9\%$ on HPDv3. These results clearly indicate that the HPDv3 dataset provides the most comprehensive and effective supervision for training. It significantly enhances the robustness and generalization of HPSv3, underscoring its superiority over other datasets.

\subsection{Clarification on Loss Function}

In this section, we clarify that various loss functions mentioned in the literature~\cite{xu2023imagereward, kirstain2023pick, hpsv2, mps} (including this paper) and the bradley-terry loss~\cite{bradley1952rank} share the same underlying optimization objective.
\paragraph{Form 1: Optimizing KL-divergence.} In \cite{hpsv2, mps, kirstain2023pick}, the predicted preference $\hat{y_i}$ is calculated as::
\begin{equation}
\hat{y}_i=\frac{\exp \left(r_i\right)}{\sum_{j=1}^2 \exp \left(r_j)\right)},
\end{equation}
where $r_i$ denotes the preference score of sample $x_i$. And the model is optimized by minimizing the KL-divergence between the ground truth $y$ and the predicted distribution. Specifically, $y=[1,0]$ if sample $x_1$ is preferred over $x_2$, and $y = [0, 1]$ otherwise.
The loss function is formalized as below to minimize KL-divergence:
\begin{equation}
L_{\mathrm{pref}}=\sum_{j=1}^2 y_i\left(\log y_i-\log \hat{y_j}\right) .
\end{equation}

To simplify the problem, we assume that the sample $x_h$ is the preferred sample, while $x_l$ is the dispreferred one. According to the order, $y_i$  will always be $1$. Substituting them into the loss function, we obtain:
\begin{equation}
\begin{split}
 L_{\mathrm{pref}} &= -\log \left(\frac{\exp(r_h)}{\exp(r_h) + \exp(r_l)}\right) \\
 &= -\log \left(\frac{1}{1 + \exp(r_l - r_h)}\right) \\
 &= \log \left(1 + \exp(r_l - r_h)\right).   
\end{split}
\end{equation}
This shows that the KL-divergence formulation reduces to a logistic loss comparing the preference scores of the two samples. 

\paragraph{Form 2: Bradley-Terry Loss.}
\cite{xu2023imagereward, bradley1952rank} adopt the Bradley-Terry loss to maximize the probability of the winning (higher-ranked) samples over the losing (lower-ranked) samples. The probability of the winning samples in Bradley-Terry model can be defined as:

\begin{equation}
\begin{split}
    P(x_h \succ x_l) &= \frac{\exp(r_h)}{\exp(r_h) + \exp(r_l)} \\
    &=\frac{1}{1+\exp(r_l-r_h)}\\
    &=\text{sigmoid}(r_l-r_h).
\end{split}
\end{equation}
The goal is to maximize this probability. Therefore, the model is optimized by minimizing the negative log-likelihood:
\begin{equation}
\begin{split}
    L_{\mathrm{BT}} &= -\log P(x_h \succ x_l) \\
    & = -\log(\text{sigmoid}(r_l-r_h)) \\
\end{split}
\end{equation}
However, we can continue to simplify it.
\begin{equation}
\begin{split}
    L_{\mathrm{BT}} &= -\log(\text{sigmoid}(r_l-r_h)) \\
    &= - \log(\frac{1}{1+\exp(r_l-r_h)}) = \log \left(1 + \exp(r_l - r_h)\right).  
\end{split}
\end{equation}
\paragraph{Equivalence Conclusion.} We can observe that both Form 1 (KL-divergence) and Form 2 (Bradley-Terry) ultimately converge to the same pair-wise logistic ranking loss:
\begin{equation}
L = \log \left(1 + \exp(r_l - r_h)\right),
\end{equation}
demonstrating their fundamental equivalence in optimization objectives despite different theoretical origins.

\section{\dname Dataset Visualization}
\subsection{Dataset Visualization}

\begin{figure*}[ht]
  \centering
  \small
  \begin{subfigure}{\linewidth}
    \centering
    \includegraphics[width=0.8\linewidth]{figure/datasetvisual_1.pdf}
    \vspace{-0.1cm}
  \end{subfigure}
  
  \vspace{0.1cm}
  \begin{subfigure}{\linewidth}
    \centering
    \includegraphics[width=0.8\linewidth]{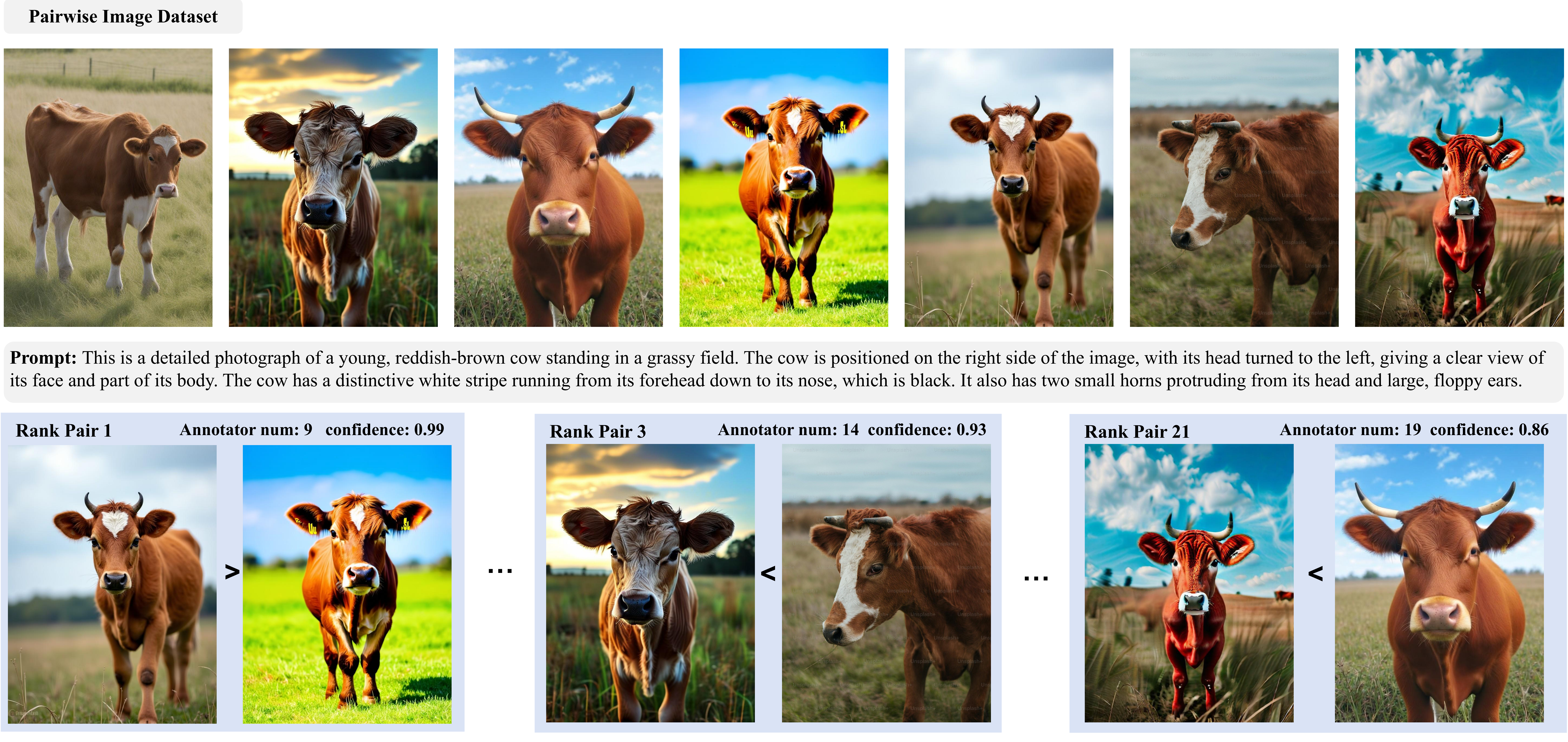}
    \vspace{-0.1cm}
  \end{subfigure}
  
  \vspace{0.1cm}
  \begin{subfigure}{\linewidth}
    \centering
    \includegraphics[width=0.8\linewidth]{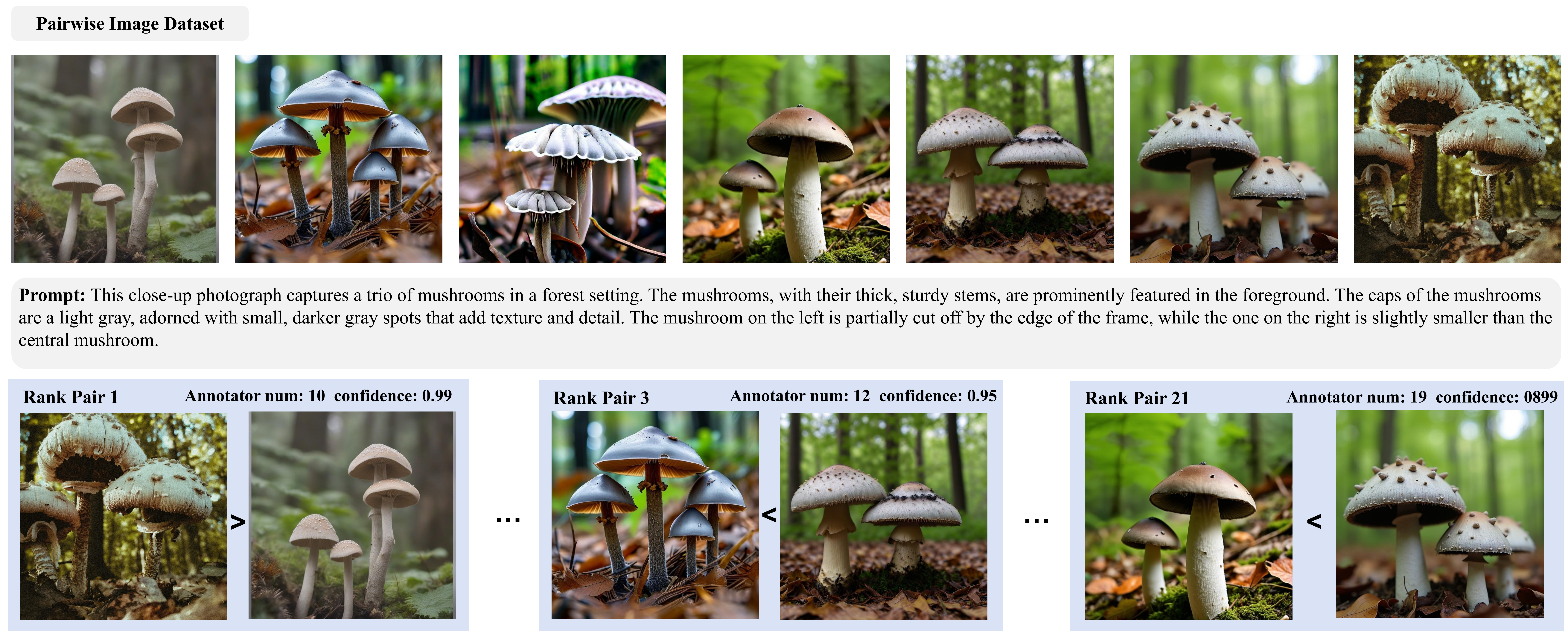}
  \end{subfigure}
  
  \caption{\textbf{\dname Dataset Visualization.} Our dataset contains a diverse range of images spanning multiple categories including animals, architecture, characters, and other subjects. Each row displays different samples demonstrating the variety and quality of the dataset.}
  \label{fig:hpdv2datasetvisual}
\end{figure*}

\Cref{fig:hpdv2datasetvisual} showcases examples from the HPDv3 dataset. Each image pair consists of different images generated from the same prompt, with the images sourced from various image generation models as well as real-world photographs. For each prompt, we systematically create pairwise comparisons by pairing all possible image combinations. Human annotators then evaluate these pairs to provide preference judgments, resulting in a detailed collection of pairwise preference data.

This rigorous pairwise annotation approach captures nuanced human preferences across different visual representations of the same concept. By including both AI-generated images from diverse models and real photographs, the dataset enables a comprehensive analysis of human preferences across the spectrum of synthetic and authentic visual content. The resulting preference signals serve as a rich foundation for training our HPSv3 model to better align with human judgments.

\subsection{Benchmark Visualization}
\Cref{fig:hpdv3_prompts}, \ref{fig:hpdv3_prompts1} and \ref{fig:hpdv3_prompts2} show sample prompts from the HPDv3 Benchmark.  This benchmark is designed as a diverse and standardized evaluation framework for assessing the performance of image generation models. Specifically, the HPDv3 Benchmark includes a carefully curated set of $1,000$ prompts for each of the $12$ categories, drawn from three datasets: HPDv3, HPDv2, and JourneyDB. These prompts cover a variety of styles and lengths to ensure comprehensive evaluation.

For prompts sourced from the HPDv3 dataset, we include the corresponding real-world reference images to facilitate comparisons with generated images. On the other hand, prompts from the HPDv2 and JourneyDB datasets are provided as text-only, focusing on evaluating a model’s ability to generate images purely from textual input.

This setup enables evaluation from multiple perspectives. The inclusion of real-world reference images provides a way to measure how closely generated images align with actual visuals. At the same time, the text-only prompts test the model's ability to interpret and generate images solely based on textual descriptions. By combining these approaches, the HPDv3 Benchmark offers a comprehensive framework to assess image synthesis quality across various content categories and prompt styles, promoting consistent evaluation and progress in text-to-image generation research.

\section{More Result of CoHP}
\begin{figure*}[t]
  \centering
  \small
  \includegraphics[width=0.8\linewidth]{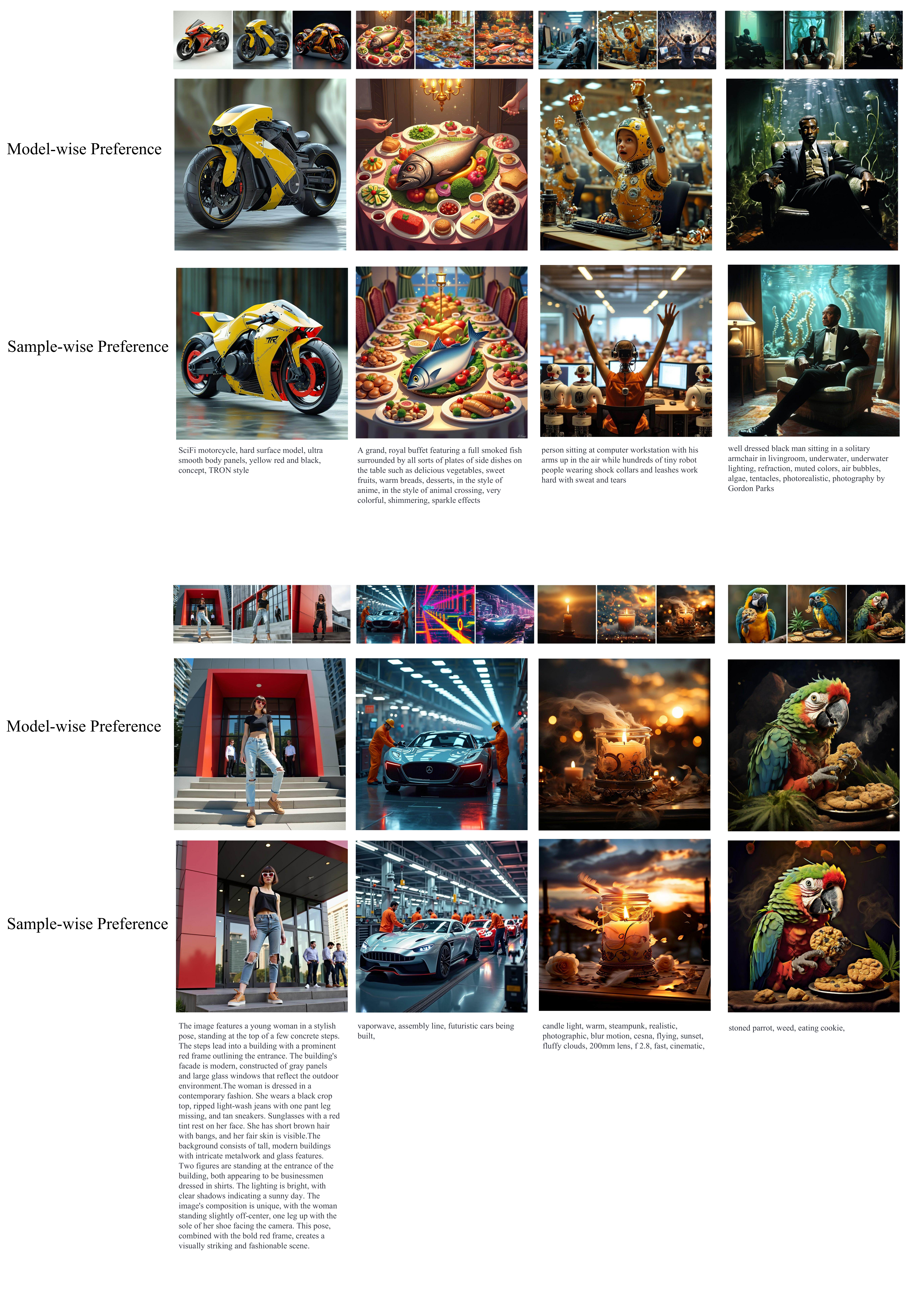}
  \caption{\textbf{More Result of CoHP}}
  \label{fig:cohp_more_result1}
\end{figure*}

\begin{figure*}[t]
  \centering
  \small
  \includegraphics[width=0.8\linewidth]{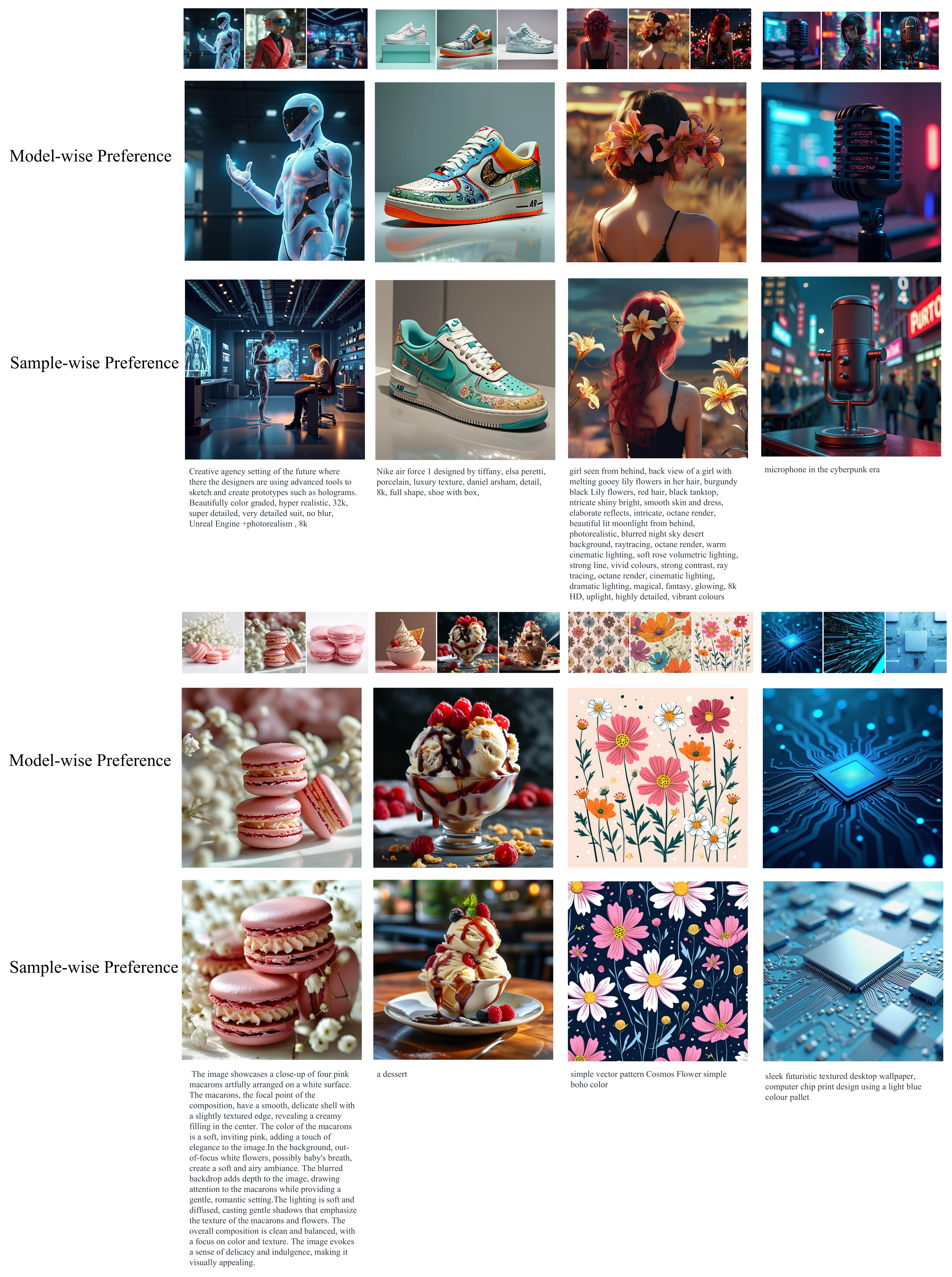}
  \caption{\textbf{More Result of CoHP}}
  \label{fig:cohp_more_result2}
\end{figure*}

In this section, we present an extensive collection of generation results from CoHP. We showcase diverse outputs produced across multiple iterations. The first row of \Cref{fig:cohp_more_result1} and \ref{fig:cohp_more_result2} shows the best result of each model (Flux, Kolors and Playground v2.5) generated in the Model-wise preference stage. As illustrated in \Cref{fig:cohp_more_result1} and \Cref{fig:cohp_more_result2}, the Model-wise preference stage plays a critical role in CoHP by selecting the best model that can generate images with strong semantic understanding and well-constructed compositions. Meanwhile, the sample-wise preference stage contributes by refining the images and enhancing their details.

\begin{figure*}[t]
  \centering
  \includegraphics[width=\linewidth]{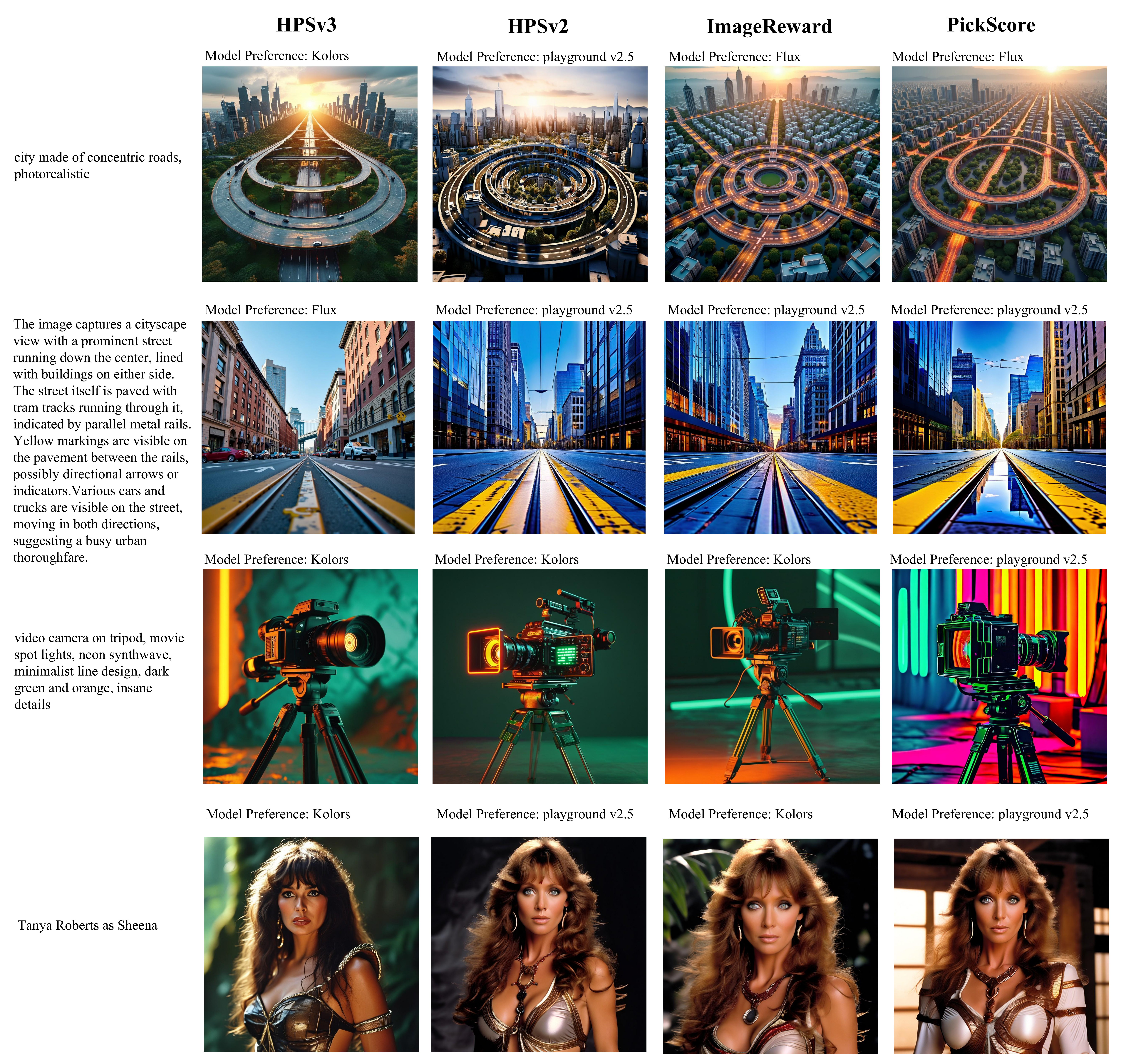}
  \caption{\textbf{More Result of CoHP with different human preference models}}
  \label{fig:cohp_compare_more}
\end{figure*}

\begin{figure*}[t]
  \centering
  \includegraphics[width=\linewidth]{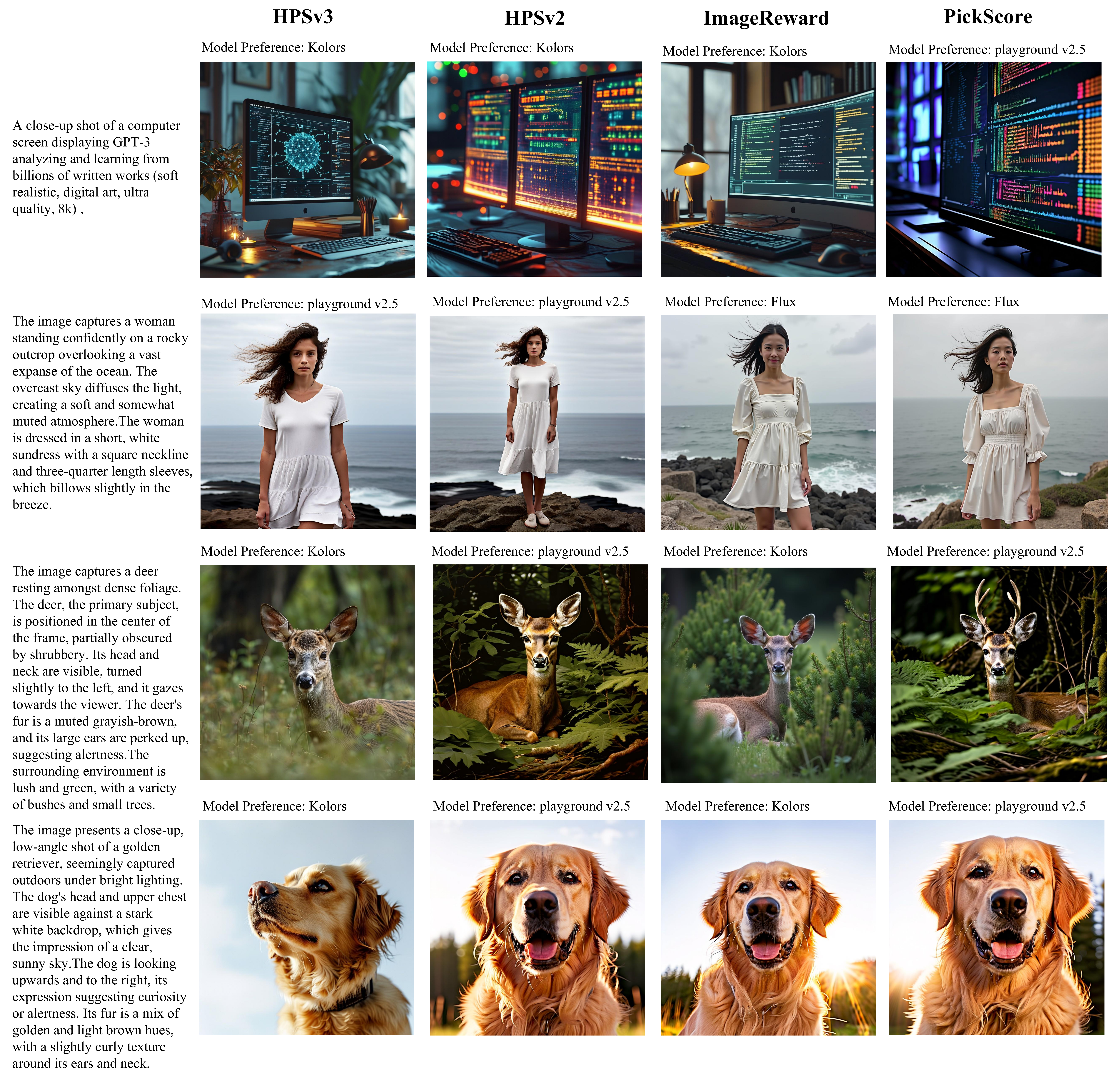}
  \caption{\textbf{More Result of CoHP with different human preference models}}
  \label{fig:cohp_compare_more1}
\end{figure*}

\Cref{fig:cohp_compare_more} and \ref{fig:cohp_compare_more1} demonstrate comparative results obtained by implementing various human preference models within our framework. These expanded visualizations provide insights into how different preference modeling approaches influence the quality, diversity, and human-alignment of the generated images. 

\section{HPSv3 as Reward Model}
When using reinforcement learning (RL) to improve the quality of generated images, the design of the reward model is critically important. A well-designed reward model can significantly improve outputs by boosting realism, aesthetic quality, and text-image alignment, or by aligning outputs more closely with human preferences. Leveraging a carefully built wide-spectrum image quality dataset and a backbone based on a Visual Language Model, HPSv3 excels at capturing human preferences more accurately. This reduces reward hacking behaviors in RL, guiding the model to produce content that better matches human expectations.

\noindent\textbf{DanceGRPO} We employ DanceGRPO \cite{xue2025dancegrpo} as the reinforcement learning algorithm for image generation and compare the results when using ImageReward, PiscScore, HPSv2 and HPSv3 as reward models, respectively. DanceGRPO performs multiple sampling of diffusion trajectories, scores the final generated image of each trajectory using the reward model, and conducts policy gradient optimization by calculating the advantage value of each trajectory relative to the average reward, thereby improving the model's performance. For all reward models, we use the same default experimental settings in DanceGRPO. We use Stable-Diffusion v1.4\cite{Rombach_2022_CVPR} as our base model, and performs 300 training iteration.

\begin{figure*}[t]
  \centering
  \includegraphics[width=0.8\linewidth]{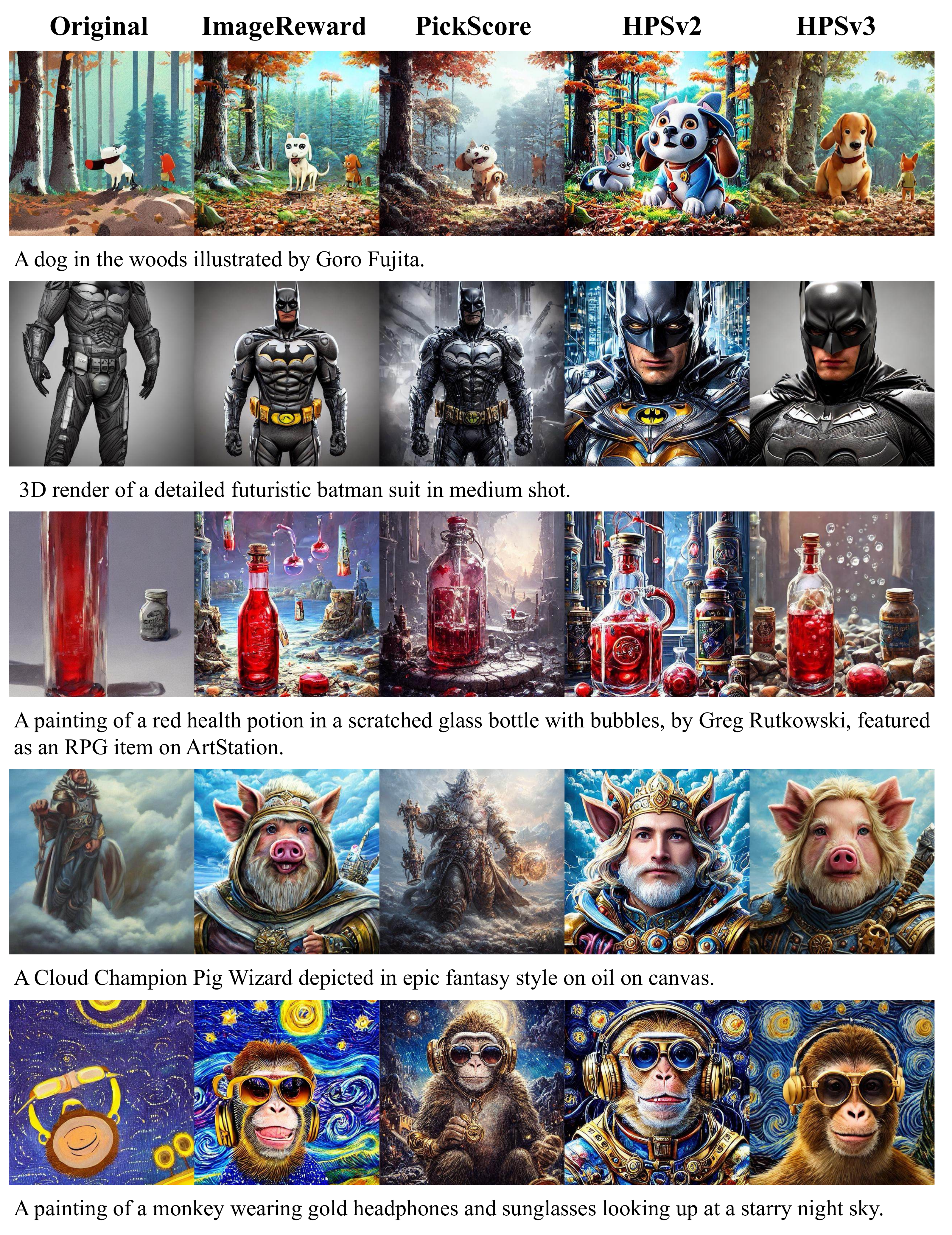}
  \caption{\textbf{Results of DanceGRPO Using Different human preference models}}
  \label{fig:RL_result}
\end{figure*}

\begin{figure*}[t]
  \centering
  \includegraphics[width=0.8\linewidth]{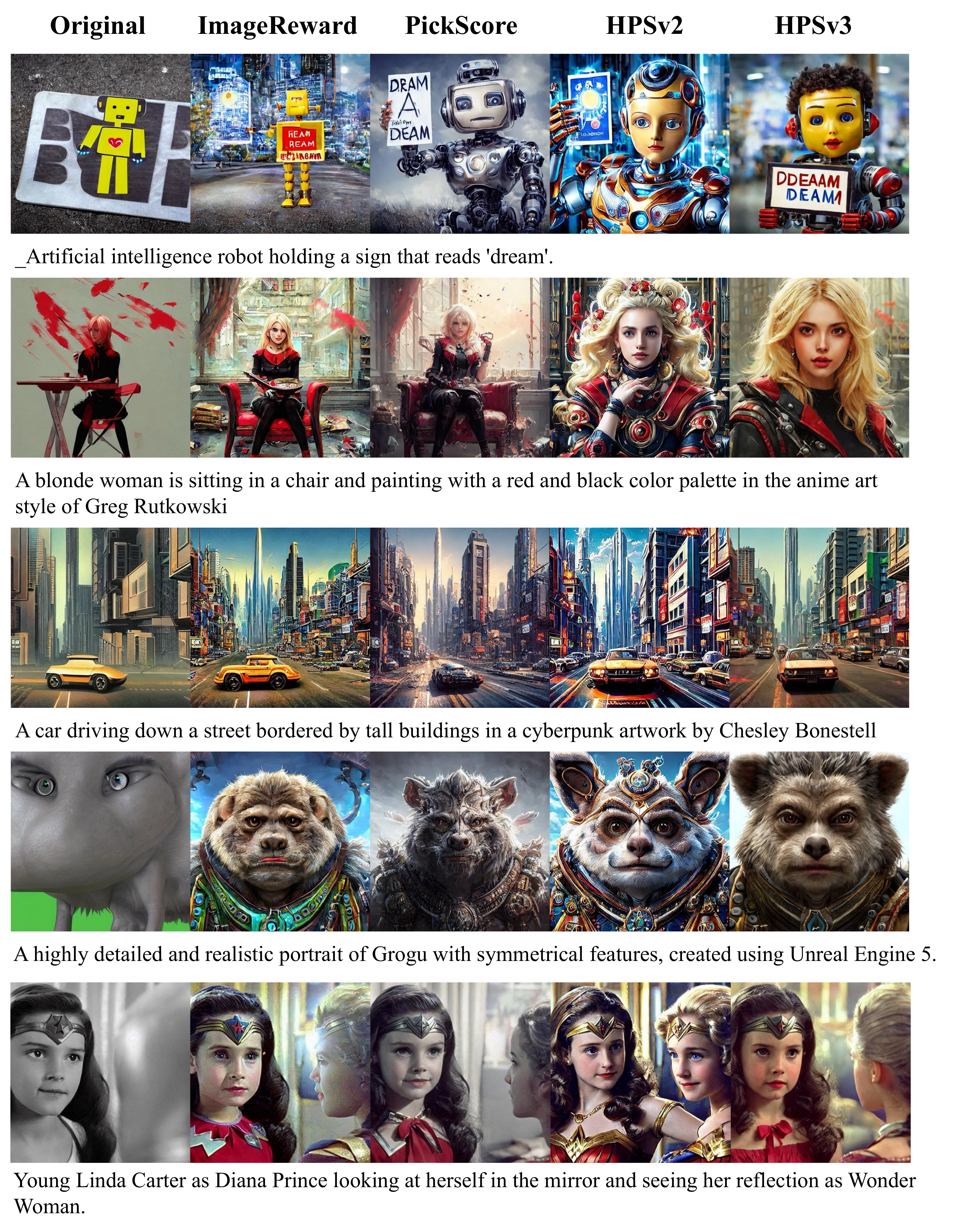}
  \caption{\textbf{Results of DanceGRPO Using Different human preference models}}
  \label{fig:RL_result2}
\end{figure*}

\begin{figure*}[t]
  \centering
  \includegraphics[width=0.9\linewidth]{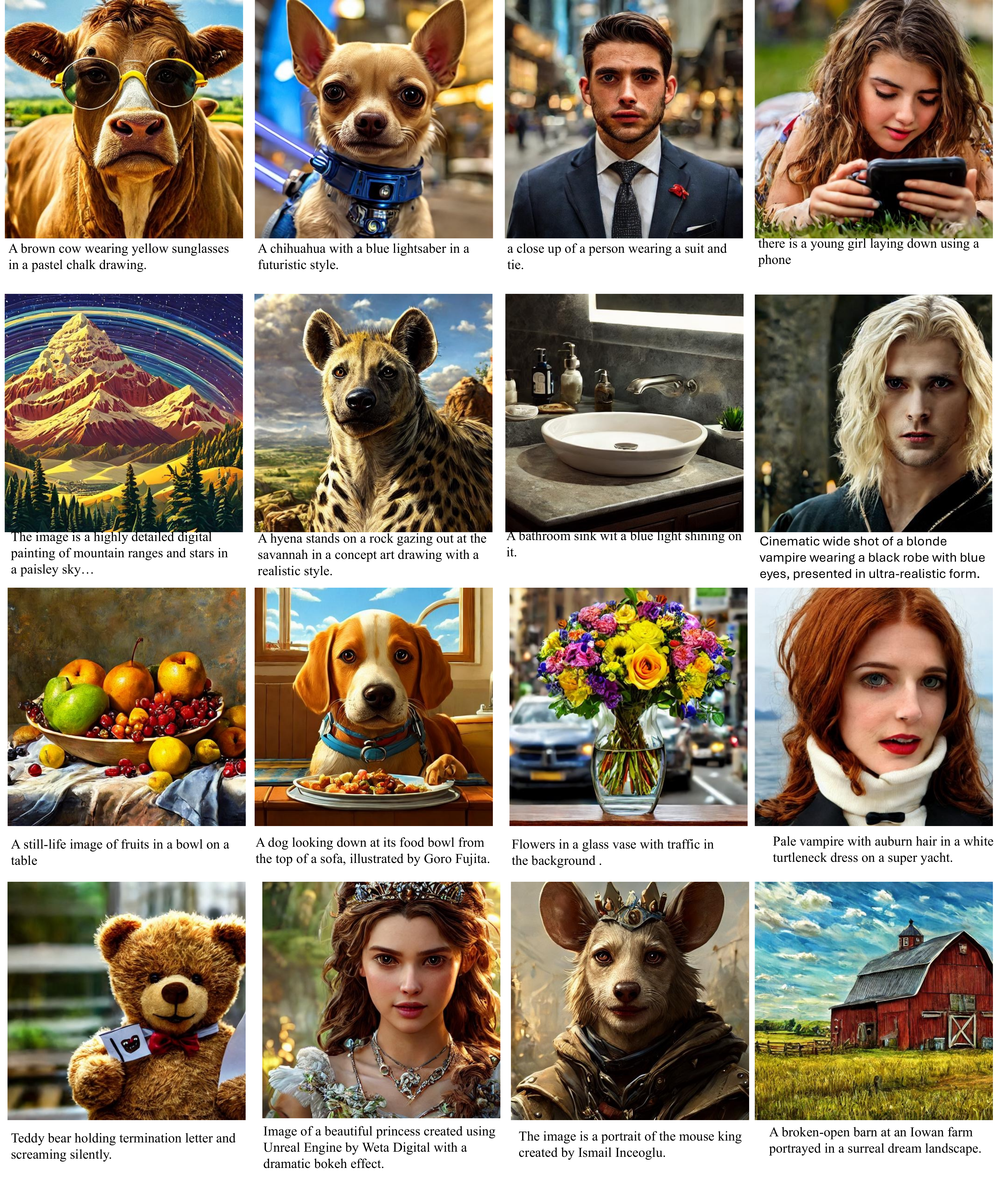}
  \caption{\textbf{More Results of DanceGRPO Using HPSv3 as Reward Model Based on SD1.4}}
  \label{fig:RL_result3}
\end{figure*}

\noindent\textbf{Experiment Results} \Cref{fig:RL_result,fig:RL_result2} presents qualitative results obtained after the same number of training iterations. For convenience, we refer to the image generation models trained with these reward models as $M_{\text{ImageReward}}$, $M_{\text{Pickscore}}$, $M_{\text{HPSv2}}$ and $M_{\text{HPSv3}}$, respectively. The results show that all these reward models improve the quality and aesthetic appeal of the generated images. $M_{\text{HPSv3}}$ produces images with greater realism—more natural color saturation, smoother lighting and shadows, and fewer artifacts and distortions. Moreover, $M_{\text{HPSv3}}$ exhibits less reward hacking. As shown in the first column of the third row, the second column of the fifth row, and the first and second columns of the sixth row, $M_{\text{HPSv2}}$ tends to generate many meaningless accessories, objects not mentioned in the prompt, or decorative light effects and spots. This behavior suggests that the model is engaging in reward hacking through these elements, whereas $M_{\text{HPSv3}}$ exhibits significantly less of this phenomenon. More results of DanceGRPO using HPSv3 as the reward model are shown in \Cref{fig:RL_result3}.

\section{Term of Use of HPDv3}

\noindent\textbf{Ownership and Responsibility.}
The HPDv3 dataset contains some parts of images obtained from the Internet, which are not the property of MizzenAI. MizzenAI is not responsible for the content or the meaning of these images.

\noindent\textbf{Noncommercial Usage.}
Our funding resources, dataset, and models are strictly limited to noncommercial use. This aligns with the principle of "fair use" as suggested by the United States Supreme Court for educational and research purposes. Any use of the HPDv3 dataset for commercial purposes is strictly prohibited.

\noindent\textbf{Restrictions on Usage.}
You agree not to reproduce, duplicate, copy, sell, trade, resell, or exploit, for any commercial purposes, any portion of the images or any portion of derived data from the HPDv3 dataset. You also agree not to further copy, publish, or distribute any portion of the HPDv3 dataset. However, it is permitted to make copies of the dataset for internal use at a single site within the same organization.

\noindent\textbf{Removal of Content.} If you wish to have your content or product removed from the HPDv3 dataset, please contact us, and we will address your request promptly.

\noindent\textbf{Acceptance of Terms.}
By using the HPDv3 dataset, you agree to comply with these Terms of Usage. Any violation of these terms may result in the termination of your access to the dataset and may lead to legal action.

\noindent\textbf{Licensing Policy.}
To prevent unauthorized commercial usage of our dataset and models, we employ the "CC BY-NC-SA" license (Creative Commons Attribution-NonCommercial-ShareAlike). This license permits others to freely share and adapt our work for non-commercial purposes, provided proper attribution is given, and derivative works maintain the same licensing terms. This measure ensures ethical distribution while preserving our original intent for non-commercial use.

\noindent\textbf{Image Collection and Licensing Compliance.}
A significant portion of the real images in our dataset are sourced from Unsplash, a platform offering high-quality images through the CC0 license. The CC0 license permits unrestricted collection, distribution, and use of images, including free utilization for research and educational purposes. By inheriting these licensing terms, we ensure compliance with intellectual property standards while fostering free and open collaboration within the research community.

\noindent\textbf{Commitment to Ethical and Fair Usage.}
We are committed to maintaining ethical standards in dataset construction and model development. All materials have been vetted to ensure adherence to licensing agreements and proper attribution where applicable. We encourage the broader community to uphold these ethical principles when utilizing our work, thereby fostering responsible research practices and avoiding any misuse of intellectual property.

\section{Limitation}

While HPDv3 contains 1.08M text-image pairs and 1.17M pairwise data, aiming to reflect real-world user preferences, it is important to acknowledge its inherent limitations, which may affect its generalizability and applicability in certain contexts.

\noindent\textbf{Prompt Distribution Bias}
The dataset construction is primarily based on the prompt categories frequently observed in the JourneyDB database, which reflects general user input patterns. While this approach captures a broad range of typical generative use cases, it may inadvertently overlook specialized domains such as medicine, biology, physics, and other specialized fields requiring unique data. For example, generative models tailored for medical imaging or scientific diagram generation might not perform accurately when benchmarked with our dataset. This potential bias could limit the dataset's usefulness for evaluating models designed for these specialized applications.

\noindent\textbf{Unified Scoring Metric}
Our annotation pipeline employs a unified scoring metric to evaluate text-image pairs holistically across all dimensions. While this approach simplifies the evaluation process, it does not provide insights into more fine-grained dimensions such as color fidelity, artistic style, sharpness, or image clarity. This lack of granularity might hinder more detailed analysis and benchmarking of generative models, especially for applications where specific attributes are critical.

\noindent\textbf{Annotator Demographics}
The dataset annotation process did not enforce strict demographic controls or categorizations. Information about annotators' ethnicity, age, professional expertise, and cultural background was not collected or utilized during data annotation. As a result, the annotations may reflect unintended bias based on the subjective perspectives of the annotators. This lack of demographic diversity could reduce the robustness of the dataset in evaluating generative models designed for global or culturally sensitive contexts.

\noindent\textbf{Challenges in Difficult Cases}
To ensure robust annotations, we adopted a multi-annotator approach for labeling text-image pairs, allowing feedback from multiple individuals to improve the reliability of scores. However, this approach encountered challenges when dealing with difficult or ambiguous cases. For prompts or images that were subjective or had conflicting interpretations, annotators often struggled to converge on a consistent score. These unresolved discrepancies can affect the accuracy of the dataset and limit its ability to serve as a definitive benchmark in such cases. Despite these challenges, the multi-annotator mechanism remains a valuable method for improving dataset reliability overall.

\input{tables/supp_benchmark_prompt}

%% file: tables/sourcemodel.tex
\begin{table*}[t]
   \centering
   \small
   
    \setlength\tabcolsep{0.9mm}{
    \scalebox{1.0}{
    \begin{tabular}{lllll}    
    \toprule
     \textbf{Image Source} & \textbf{Type} & \textbf{Num Images} & \textbf{Prompt Source} & \textbf{Split} \\
     \midrule
     High Quality Image (HQI) & Real Image & 57759 & VLM Caption & Train \& Test \\
     Midjourney & - & 331955 & User Input & Train \\
     CogView4~\cite{cogview3} & DiT & 400 & HQI+HPDv2+JourneyDB & Test \\ 
     FLUX.1 dev~\cite{flux2024} & DiT & 48927 & HQI+HPDv2+JourneyDB & Train \& Test \\ 
     Kolors~\cite{kolors} & DiT & 49705 & HQI+HPDv2+JourneyDB & Train \& Test \\ 
     HunyuanDiT~\cite{hunyuandit} & DiT & 46133 & HQI+HPDv2+JourneyDB & Train \& Test \\ 
     Stable Diffusion 3 Medium~\cite{sd3} & DiT & 49266 & HQI+HPDv2+JourneyDB & Train \& Test \\ 
     Stable Diffusion XL~\cite{podell2023sdxl} & Diffusion & 49025 & HQI+HPDv2+JourneyDB & Train \& Test \\ 
     PixArt-$\Sigma$~\cite{pixart} & Diffusion & 400 & HQI+HPDv2+JourneyDB & Test \\ 
     Infinity~\cite{han2024infinity} & Autoregressive & 27061 & HQI+JourneyDB & Train \& Test \\ 
     Stable Diffusion 2~\cite{stable_diffusion} & Diffusion & 19124 & HQI+JourneyDB & Train \& Test \\ 
     CogView2~\cite{ding2022cogview2} & Autoregressive & 3823 & HQI+JourneyDB & Train \& Test \\ 
     FuseDream~\cite{liu2021fusedream} & Diffusion & 468 & HQI+JourneyDB & Train \& Test \\ 
     VQ-Diffusion~\cite{vqgan} & Diffusion & 18837 & HQI+JourneyDB & Train \& Test \\ 
     Glide~\cite{glide} & Diffusion & 19989 & HQI+JourneyDB & Train \& Test \\ 
     Stable Diffusion 1.4~\cite{stable_diffusion} & Diffusion & 18596 & HQI+JourneyDB & Train \& Test \\ 
     Stable Diffusion 1.1~\cite{stable_diffusion} & Diffusion & 19043 & HQI+JourneyDB & Train \& Test \\ 
     HPDv2~\cite{hpsv2} & / & 327763 & - & Train \\ 
     \midrule
     \textbf{Total} &  & 1088274 &  &  \\
     \bottomrule
  \end{tabular}}}
  
 \caption{\textbf{Image Sources of \dname.} The dataset contains images from both high-quality real photographs and various types of image generation models, including autoregressive models, DiT-based models, and diffusion models.}
 \label{tab:sourcemodel}
\end{table*}

%% file: tables/supp_score_standard.tex
\begin{table}[!t]
    \centering
    \renewcommand{\arraystretch}{1.2}
    \scalebox{0.6}{\begin{tabular}{|p{14cm}|}
        \hline
        \textbf{Standard} \\ 
        \hline
        \small
        Based on the text you see two pictures, please combine the degree of detail finesse, chipping, text, artistry, aesthetics and other dimensions of comprehensive consideration, choose the one you prefer.
        
Tips: do not only consider the text and image correlation, do not consider the impact of image size, need to combine multiple dimensions to make a comprehensive judgment!
\\
        
        \hline
    \end{tabular}}
    \caption{\textbf{Details of the annotation guideline.}}
    \label{tab:score}
\end{table}

%% file: tables/supp_annotate_result.tex
\begin{table}[!t]
\centering

\begin{tabular}{|p{0.95\linewidth}|}
\hline

\textbf{Example 1: Question ID: xxxxxxx} \\
\hline
\textbf{Image Type:} Hunyuan VS SD3 \\
\textbf{Ground Truth:} 1 \\
\textbf{Confidence Score:} 0.98 \\
\textbf{Average Completion Time:} 9.56 seconds \\
\textbf{Response Distribution:} 8/9 users (88.9\%) correctly identified the image as synthetic \\
\textbf{User Capability:} Only one user had measured capability score (68.0) \\
\textbf{Speed Range:} From 1.269s to 22.814s \\
\hline
\textbf{Example 2: Question ID: xxxxxxx} \\
\hline
\textbf{Image Type:} Real image VS SD3 \\
\textbf{Ground Truth:} 0 \\
\textbf{Confidence Score:} 0.95 \\
\textbf{Average Completion Time:} 5.74 seconds \\
\textbf{Response Distribution:} 19/19 users (94.7\%) incorrectly identified the real image as synthetic \\
\textbf{User Capability:} Three users had measured capability scores (68.0, 68.0, 73.5) \\
\textbf{Speed Range:} From 0.945s to 22.253s \\
\hline
\end{tabular}
\caption{User response analysis for pairwise image analysis tasks}
\label{tab:annotation_result}
\end{table}

%% file: tables/train_data.tex
\begin{table}[!t]
\centering
\small

\scalebox{0.8}{
\begin{tabular}{lccc}
\toprule
\textbf{Data Source} & \textbf{Description} & \textbf{Pair Num}  \\ 
\midrule
\multirow{2}{*}{HPDv3}  & Real images and comparisons & 652k  \\ 
                        & Golden trainset filtered from HPDv2 & 250k  \\ 
\midrule
Pick-A-Pic             & Randomly selected subset & 350k  \\
ImageReward            & Randomly selected subset & 120k \\
\midrule
Midjourney             & Real user choice data & 150k \\ 
\midrule
\textbf{Total}         & \textbf{} & \textbf{1522k} \\ 
\bottomrule
\end{tabular}}
\caption{\textbf{Composition of the training dataset used for HPSv3 model training.}}
\label{tab:training_data}
\end{table}

%% file: tables/datasetablation.tex
\begin{table}[!t]

  \centering
  \setlength\tabcolsep{0.9mm}{
  \scalebox{0.8}{
  \begin{tabular}{llcccccc}
    \toprule
    Model & ImageReward & PickScore & HPDv2 & HPDv3 \\
    \midrule
    HPSv3 (HPDv2) & 62.6 & 64.4 & 82.3 & 66.6 \\
    HPSv3 (ImageReward) & 65.5 & 64.4 & 80.5 & 63.4\\
    HPSv3 (PickScore) & 61.0 & 70.6 &  80.2 & 64.9\\
    HPSv3 (Ours) & \textbf{66.8} & \textbf{72.8} & \textbf{85.4} & \textbf{76.9}\\
    \bottomrule
  \end{tabular}
  }}
\caption{\textbf{Dataset ablation}. We train HPSv3 using the training datasets from HPDv2, ImageReward, and PickScore. The results demonstrate that training with HPDv3 training dataset achieves the highest accuracy across all test sets, showcasing its superior performance.}
\label{tab:datasetablation}
\end{table}

%% file: tables/supp_benchmark_prompt.tex
\begin{figure*}[t]
\centering
\fbox{
\begin{minipage}{0.9\textwidth}

\textbf{Animals:}
\begin{itemize}
\item water color Bird similar to alex grey style , opal.
\item an armadillo on a bicycle in the rain.
\item American opossum in faerieland, cute, in the style of Maurice Sendak.
\item The image showcases a magnificent peacock with its tail fully fanned out, displaying its vibrant plumage. The bird's body is a rich, deep blue, contrasting beautifully with the array of colors in its feathers. The head is adorned with a small crest of feathers, and a white marking extends from the beak along the side of the face.The tail feathers are the focal point, featuring iridescent eye patterns in shades of blue, green, and gold. Each eye is surrounded by a halo of darker hues, creating a mesmerizing effect. ...

\end{itemize}

\textbf{Architecture:}
\begin{itemize}
\item beautiful atmospheric picture of ghosts attacking New York, Y\u014dkai, visually stunning, highly detailed, 8K,
\item fantasy Christmas house, in a field of snow, fairy lights, by Gediminas Pranckevicius
\item the palace of the Red Branch Knights
\item The image captures a row of Tudor-style buildings under a cloudy sky. The buildings are characterized by their distinctive black and white timber framing, a hallmark of Tudor architecture. The black beams create vertical and diagonal patterns against the white-painted walls, giving the facades a striking contrast and a sense of depth. The roofs are steeply pitched and covered with dark tiles.

\end{itemize}

\textbf{Arts:}
\begin{itemize}
\item elf martial artist, meditating, misty background, pink flowers on the ground, fantasy art, oil painting
\item dreamy watercolor painting of an angry witch and her ravens in a magical forest
\item sketch, Abstract Purple background, illustration, watercolor::2.5,
\item The image captures a tourist immersed in art appreciation within a museum setting. The focus is on the back of a young man with light-colored hair, wearing a white t-shirt, dark jeans, and a black backpack. He stands facing a display of classical sculptures.The sculptures include a prominent male nude figure with its arm raised, positioned to the left of a smaller statue. ...

\end{itemize}

\textbf{Characters:}
\begin{itemize}
\item The image features a half cyborg girl character design with a nature meets technology theme, rendered with cinematic lighting and high detail.
\item Anime girl in transparent holographic light suit black and yellow, Full body, three quarter length + poster + Guweiz + Cyberpunk, branding, texts, labels, three quarter body
\item the most beautiful Andorran woman in the world,
\item The image features a medium shot of a man with dark skin, his gaze directed straight at the viewer. He has a serious expression, with a neatly trimmed goatee and dark eyes that carry a piercing intensity. His hair is styled in thin braids, some falling around his face and shoulders, adding to his thoughtful appearance. ...

\end{itemize}

\textbf{Design:}
\begin{itemize}
\item illustration of bluebeard tale, with typographic placement , linocut, by Saul Bass
\item A pen and copic marker sketch of the design of an simple robotic fish, white background, no paper background, no words, no pen or marker in photo, high quality
\item vector symbol Sci fi Space 70s empty world
\item This image shows a stylish and monochromatic interior design. The walls are painted in a deep, matte black, which serves as a dramatic backdrop for the decor. A black upholstered armchair with button detailing and four wooden legs sits prominently in the foreground, adding a touch of luxury. ...

\end{itemize}
\end{minipage}}
\caption{Representative examples of prompts from the HPDv3 Benchmark. For each category, we include a range of prompts varying from simple descriptions to highly detailed specifications. These prompts are used to generate image pairs from different AI models for calculating \mname Score of each model.}
\label{fig:hpdv3_prompts}
\end{figure*}

\begin{figure*}[t]
\centering
\fbox{
\begin{minipage}{0.9\textwidth}

\textbf{Food:}
\begin{itemize}
\item a handful of guava fruits drawn by david hockney, detailed, intricate
\item sketch round cracker with chocolate Half enrobbed with color
\item cartoon wine illustration, vector, simple clean, minimalist, wallpaper, bright, collection, in a set
\item The image displays two tall glasses filled with a deep red beverage, likely a type of cocktail or juice, set against a dark background. The glasses are garnished with slices of grapefruit and sprigs of fresh mint, adding a vibrant splash of color and freshness. Ice cubes float within the drink, suggesting a refreshing and chilled experience.The wooden surface beneath the glasses has a weathered, rustic look, enhancing the natural and organic feel of the composition. One glass is positioned closer to the viewer, bringing the details of the drink and garnish into sharp focus, while the other is slightly blurred in the background, creating depth. ...

\end{itemize}

\textbf{Natural Scenery:}
\begin{itemize}
\item an epic outdoor sticker of the animas river with the san juan mountains in the background
\item water ocean texture shot from bird perspective
\item the view from inside a cosmic black hole 8k photorealistic
\item The image presents a long, straight stretch of asphalt road leading towards a distant mountain range. The road takes up most of the foreground, its surface a dark gray, with two solid yellow lines running down the center, creating a strong sense of perspective. On either side of the road, there are dark, undefined areas of land.In the distance, snow-capped mountains dominate the horizon. Their jagged peaks and white surfaces contrast sharply with the dark land and road below, adding depth and a sense of grandeur to the scene.The sky above is partially cloudy, with patches of blue peeking through the white clouds. The clouds are scattered across the sky, adding texture and visual interest to the upper part of the image. ...

\end{itemize}

\textbf{Plants:}
\begin{itemize}
\item fotorealistic jungle leaves, repetitive pattern, endless, mid-century modern style, geometric
\item dark geisha pink flower with sword blue background ultra hd 8k
\item garden full of golden flowers,
\item The image presents a vast, golden field of ripe wheat under a clear blue sky. The wheat stalks dominate the foreground, their heads heavy with grain, creating a dense and textured visual tapestry. The color palette is dominated by warm tones, with the golden wheat contrasting nicely with the cool blue above.The wheat field stretches far into the distance, appearing to meet a slightly darker horizon line. A few scattered white clouds add a touch of lightness to the sky. The composition is simple yet striking, emphasizing the expansive nature of the landscape and the abundance of the harvest. The lighting is bright and sunny, casting gentle shadows within the field and highlighting the individual grains of wheat. Overall, the image evokes a sense of warmth, tranquility, and the bounty of nature. ...

\end{itemize}

\textbf{Products:}
\begin{itemize}
\item simple knolling, snow removal Snowblower and graders and loaders, white background
\item Hand holding megaphone on bright yellow background with plenty of copy space. Magazine collage cut out style
\item raw wrapping papers inspired Yeezy 350, hyper-detailed
\item The image presents a striking contrast between black and red, with a small, wrapped gift placed on the black portion of the background. The background is divided diagonally, with a textured black area on the left and a smooth, vibrant red area on the right. This division creates a visually compelling composition.The gift itself is small and square, wrapped in shiny red foil. A bright red ribbon is tied around the package in a simple bow, with the ends of the ribbon elegantly trailing onto the black background. ...

\end{itemize}

\vspace{0.3cm}
\end{minipage}}
\caption{Representative examples of prompts from the HPDv3 Benchmark. For each category, we include a range of prompts varying from simple descriptions to highly detailed specifications. These prompts are used to generate image pairs from different AI models for calculating \mname Score of each model.}
\label{fig:hpdv3_prompts1}
\end{figure*}

\begin{figure*}[t]
\centering
\fbox{
\begin{minipage}{0.9\textwidth}

\textbf{Transportation:}
\begin{itemize}
\item Sports Bike, race track, realism, 4K, No logo, Ducati, HDR, ar 3:2
\item Porsche made from candy, beautiful, editorial photography, color graded, masterpiece
\item sunken pirate ship, cinematic lighting, 4k, 8k, unreal engine, octane render
\item The image showcases a collection of vintage automobiles, arranged neatly on a paved surface that seems to be a brick-patterned area. The backdrop includes lush greenery of trees and a unique architectural structure, possibly a modern park feature with a curved design, suggesting a blend of historical vehicles in a contemporary setting.Dominating the view are three cars of different makes. The first one, partially visible on the left, is a sleek, dark green, with a closed top. Adjacent to it is a classic black car with an open driver's area, a notable upright grille, and what looks like a soft top. Positioned in the foreground is a lighter-colored vehicle, likely a light grey or off-white, also with a soft top. It features spoked wheels and shiny headlamps. All three vehicles exhibit signs of age and careful preservation. The black and grey cars each have a sign or placard visible in the front window. Overall, the image captures a timeless elegance. ...

\end{itemize}

\textbf{Science:}
\begin{itemize}
\item a solar system view of a large space battle set in the Star Wars Universe, hyper realistic, 4k resolution
\item 3D illustration of employees working in factory, 21st century
\item futuristic neuronal network with robotic integration. Black background, 8K, hyperrealistic.
\item reactor round underground scifi, hardsurface, HD, cinematography, low viewpoint, photorealistic, epic composition, Cinematic, Color Grading, portrait Photography, Ultra-Wide Angle, hyper-detailed, beautifully color-coded, insane details, intricate details, beautifully color graded, Unreal Engine, Cinematic, Color Grading, Editorial Photography, Photography, Photoshoot, Depth of Field, DOF, Tilt Blur, White Balance, 32k, Super-Resolution, Megapixel, ProPhoto RGB, VR, Halfrear Lighting, Backlight, Natural Lighting, Incandescent, Optical Fiber, Moody Lighting, Cinematic Lighting, Studio Lighting, Soft Lighting, Volumetric, Contre-Jour, Beautiful Lighting, Accent Lighting, Global Illumination, Screen Space Global Illumination, Ray Tracing Global Illumination, Optics, Scattering, Glowing, Shadows, Rough, Shimmering, Ray Tracing Reflections, Lumen Reflections, Screen Space Reflections, Diffraction Grading, Chromatic Aberration, GB Displacement, Scan Lines,...

\end{itemize}

\textbf{Others:}
\begin{itemize}
\item realistic eyeball, pupil replaced with heart, up close, red, 1970s documentary photographs 35mm
\item A memorial candle is lit on a table in a dark room close up vivid colours
\item The image captures the vibrant energy of a live music festival. The foreground is filled with a large crowd, their heads a sea of diverse hairstyles and colors. Many are looking towards a massive stage structure, which dominates the left side of the frame.The stage is a complex scaffolding of metal and screens. The screens are illuminated with bright blue light, suggesting some sort of visual display accompanying the performance. A silhouette of a performer can be seen on stage, adding to the sense of a live event.Above the stage, the sky is a patchwork of white clouds against a bright blue background, creating a sense of expansive space. In the distance, the sun appears to be setting, casting a warm glow over the scene. To the right, trees in the background mark the natural setting of the festival. The overall composition captures the excitement and scale of a large-scale music event. ...
\item The image showcases a single, small, earthen oil lamp, known as a diya, resting on a reflective surface. The diya is dark brown and has a rounded shape, typical of traditional Indian oil lamps. Atop the diya, a small flame flickers, providing the main source of light in the image.The backdrop is a soft, blurred yellow-orange gradient, creating a warm and cozy atmosphere. The surface on which the diya sits is dark, shiny, and reflective, mirroring the light from the flame. This reflection adds depth and dimension to the image, amplifying the glow and creating a sense of tranquility. ...

\end{itemize}

\vspace{0.3cm}
\end{minipage}}
\caption{Representative examples of prompts from the HPDv3 Benchmark. For each category, we include a range of prompts varying from simple descriptions to highly detailed specifications. These prompts are used to generate image pairs from different AI models for calculating \mname Score of each model.}
\label{fig:hpdv3_prompts2}
\end{figure*}